\documentclass[pdflatex,sn-mathphys-num]{sn-jnl}

\usepackage[utf8]{inputenc}
\usepackage[T1]{fontenc}
\usepackage{placeins}
\usepackage{setspace}

\usepackage{graphicx}
\usepackage{url}
\usepackage{amsmath}
\usepackage{amssymb}
\usepackage{subcaption}
\usepackage{hyperref}
\usepackage{xcolor}
\usepackage{booktabs}
\usepackage{siunitx}
\usepackage{url}
\usepackage{float}
\usepackage{comment}
\usepackage{svg}
\usepackage{float}
\usepackage{pdflscape}
\usepackage[percent]{overpic}
\title{GEM-x: Exploration-driven Materials Discovery using Gaussian Mixtures and Ensemble Learning}

\author[1]{\fnm{Shauryaraj} \sur{Jilkar}}
\author[1]{\fnm{Shoubhanik} \sur{Nath}}
\author*[1]{\fnm{Ali} \sur{Mesbah}}\email{mesbah@berkeley.edu}

\affil[1]{
\orgdiv{Department of Chemical and Biomolecular Engineering},
\orgname{University of California},
\city{Berkeley},
\state{CA 94720},
\country{USA}
}

\begin{document}

\maketitle

\begin{abstract}
GGenerative modeling offers a promising route to discovering crystalline materials beyond known databases, yet existing models often sample densely populated regions of composition–structure space, limiting novelty of the generated candidates. Here, we introduce GEM-x, a Gaussian mixture flow-matching framework that operates in a learned latent space and models the denoising velocity as a Gaussian mixture parametrized by a sparsely gated mixture-of-experts network, enabling multimodal, exploration-oriented sampling. We benchmark GEM-x against six leading generative models, including the All-atom Diffusion Transformer (ADiT), and assess the energetic plausibility of 1,000 generated candidates using r$^2$SCAN density functional theory calculations. GEM-x generates stable or metastable, unique, and novel structures at a rate of 12.3\%, compared with 9.6\% for ADiT, representing a 28\% relative improvement. This advantage is driven primarily by the generation of metastable, unique, and novel structures, for which GEM-x achieves a rate of 10.4\% versus 7.1\% for ADiT, a 46\% relative improvement. Metastable phases are particularly attractive targets for materials discovery, as they can be experimentally realizable while offering functional properties superior to those of their thermodynamically stable counterparts. The findings suggest that exploration-driven generative modeling can expand the accessible materials landscape.
\end{abstract}

\section*{ Introduction}
\paragraph{}
Metastable crystalline materials constitute a substantial and technologically important portion of experimentally accessible materials space, often exhibiting functional properties that are distinct from, or even superior to, those of their thermodynamically stable counterparts \cite{sun2016metastability,aykol2018thermodynamic}. Metastable materials occupy local minima on the free-energy landscape. Although they are not the most thermodynamically stable state of their chemical system under a given condition, kinetic barriers can hinder their transformation to more stable phases, allowing them to persist over practical timescales \cite{iupacmetastability}. The  significance and experimental accessibility of metastable phases are demonstrated by several well-established materials. Diamond, metastable relative to graphite under ambient conditions \cite{palnichenko1999diamond}, is employed in thermal management applications for high-power electronics \cite{diamond2025heatspreaders} and provides a platform for nitrogen-vacancy-based quantum sensing \cite{janitz2022diamond}. Metastable austenite enables favorable strength–ductility combinations through deformation-induced martensitic transformations in advanced high-strength steels \cite{austenite2022review}, improving crash-energy absorption in automotive components \cite{raabe2020austenite}. Similar transformation-induced strengthening mechanisms have also been demonstrated in metastable high-entropy alloys \cite{li2016metastable}. Anatase TiO$_2$, metastable relative to bulk rutile \cite{hanaor2011anatase}, is commonly employed as an electron-transport layer in perovskite solar cells \cite{fatima2024tio2} and is actively investigated for photocatalytic hydrogen production \cite{tio2hydrogen2023} and as an anode material in lithium- and sodium-ion batteries \cite{liang2022anatase}. Although thermodynamic stability is widely used as a primary screening criterion in computational materials discovery \citep{wang2021stable}, restricting the search to thermodynamic ground states risks overlooking experimentally accessible metastable phases with valuable functional properties \citep{davies2018analogy}. Systematically identifying such structures alongside stable phases  offers a promising route to broadening the pool of functional materials.

Traditional materials discovery guided by human intuition is fundamentally constrained by the limited number of candidate materials that can be identified and evaluated \cite{Zeni2025MatterGen}. The materials search space is vast and high-dimensional, spanning both discrete and continuous variables, while only a small fraction of the space corresponds to physically plausible materials \citep{xie2022}. Efficiently exploring the relevant regions of composition–structure–property landscape therefore remains a longstanding challenge in crystal structure generation \citep{park_exploration_2025}. Inverse design, which seeks structures that exhibit a target property, is inherently ill-posed due to the nonlinear, many-to-one mapping from structure to property, admitting multiple or no solutions for a given target \citep{lee2024datadriven}. Moreover, evaluating each candidate often requires computationally intensive ab initio calculations, making exhaustive search impractical. Hence, practical approaches to inverse design avoid inverting the structure–property map and instead approximate it. These approaches can be broadly categorized into sample-based optimization \citep{noh2020inverse,kadulkar2022machine} and distribution-learning methods \citep{noh2020inverse,xie2022}.

In sample-based optimization, inverse design is treated as a search problem. That is, candidate structures are iteratively proposed, evaluated against an objective function that encodes the target properties, typically computed via machine-learned surrogate models or ideally ab initio calculations, and then refined or replaced based on these evaluations. In this way, instead of learning the underlying distribution of plausible materials, these methods navigate the design space directly, using feedback from each evaluation to guide subsequent candidate proposals. Bayesian optimization, for instance,  employs a probabilistic surrogate of target properties together with an acquisition function that balances exploration and exploitation in selecting subsequent candidates \citep{boyar2025crystal_lsbo}. Reinforcement learning instead formulates structure generation as a sequential decision-making problem, optimizing policy using rewards associated with desired material properties \citep{chen2025matinvent}. Evolutionary algorithms maintain a population of candidate structures and improve the population over successive generations through selection, mutation, and recombination guided by a fitness function \citep{allahyari_coevolutionary_2020}. Although these methods have been shown to effectively steer the search towards target properties, their performance depends critically on the design of the objective function, which is often nontrivial \citep{kadulkar2022machine,shi2023multiobjective}. Materials discovery is inherently a multi-objective problem, involving trade-offs between competing target properties \citep{nath2026aiguided} and thermodynamic stability. Balancing these targets requires either choosing their relative importance in advance, which can bias the search \citep{paria2020scalarization}, or establishing the Pareto front of optimal trade-offs, which demands substantially more evaluations \citep{muller2017socemo}. Moreover, when properties of target materials are evaluated using machine-learned surrogates, the optimization may exploit model errors, favoring physically implausible structures with spuriously high predicted performance \citep{fannjiang2020autofocused}. More generally, because these methods rely on repeated assessment of candidates, their efficiency depends strongly on the cost of each evaluation \citep{kotthoff2021bayesian} and degrades as the dimensionality of the search space grows \citep{zhou2024evolutionary}.

Distribution-learning methods, also known as generative models, offer an alternative by learning a probability distribution over crystal structures from existing materials data \citep{xie2022,jiao2023diffcsp}. By capturing the statistical relationships among chemical composition, atomic arrangement, and lattice geometry, these models implicitly encode what constitutes a plausible crystal, enabling new structures to be sampled directly from the learned distribution \citep{xie2022,Zeni2025MatterGen}. When conditioned on target properties, they can further approximate the inverse mapping from property to structure, potentially reducing reliance on repeated evaluation of candidates \citep{Zeni2025MatterGen}.
Generative models for crystal structures have been developed using several architectural families. Among these, Crystal Diffusion Variational Autoencoder (CDVAE) \citep{xie2022} introduces a diffusion-based variational autoencoder for periodic material generation, combining a latent representation of crystals with a decoder that denoises atomic coordinates and atom types while respecting crystal symmetries. Crystal Structure Prediction by Joint Equivariant Diffusion (DiffCSP) \citep{jiao2023diffcsp} formulates crystal structure prediction as a joint equivariant diffusion process over fractional coordinates and lattice parameters, explicitly incorporating translation, rotation and periodic invariances. Generating Materials with Riemannian Flow Matching (FlowMM) \citep{miller2024} replaces diffusion with Riemannian flow matching on crystal manifolds, enabling more flexible and efficient transport between base distributions and stable crystal structures. UniMat \citep{yang2024} proposes a unified crystal representation for scalable diffusion-based materials generation, allowing diffusion models to be trained on larger and more chemically complex crystal datasets. Flow Matching for Material Generation with Large Language Models as Base Distributions FlowLLM \citep{sriram2024} combines a fine-tuned large language model with Riemannian flow matching, using LLM-generated crystal structures as an informed base distribution that is subsequently refined by adjusting fractional atomic coordinates and lattice parameters while keeping atomic species fixed. MatterGen \citep{Zeni2025MatterGen} introduces a diffusion-based generative model that jointly denoises atomic species, fractional coordinates, and lattice parameters, incorporating adapter-based fine-tuning and classifier-free guidance to enable the generation of stable crystal structures conditioned on desired material properties. All-atom diffusion transformers (ADiT) \citep{joshi2025allatom} introduces a unified latent diffusion transformer for both periodic materials and non-periodic molecular systems, using an all-atom autoencoder and transformer-based latent diffusion model. Because they sample directly from a learned distribution, generative models can propose candidate structures without evaluating properties at each step, substantially reducing the computational cost of candidate generation \citep{xie2022,miller2024,Zeni2025MatterGen}. However, the learned distribution is only an approximation, and generated structures must still be validated, commonly through ab initio relaxation and stability analysis, to confirm their physical and energetic plausibility \citep{miller2024,Zeni2025MatterGen}. Since only a fraction of samples are typically stable, unique, and novel, this validation can become a significant computational bottleneck \citep{miller2024,Zeni2025MatterGen}.

 Despite these advances, existing generative models struggle to balance physical plausibility with structural diversity and novelty \citep{2025arXiv250816614Y}. Because they learn from distributions of known crystal structures, they tend to favor frequently represented structural motifs, limiting exploration of underrepresented regions of composition–structure space \citep{wei2026prototype}. This limitation is especially pronounced in data-scarce chemical systems, where generating energetically plausible structures beyond well-sampled regions of the training distribution remains difficult \citep{diffusion2026unexplored}. Further, an analysis of experimentally reported inorganic crystals indicates that a substantial fraction of synthesizable materials lies above the convex hull, with the degree of accessible metastability varying considerably across chemical systems \citep{sun2016metastability,aykol2018thermodynamic}. Effective generative approaches must therefore extend beyond the training distribution while maintaining energetic plausibility, enabling the discovery of both new stable crystals and accessible metastable phases.

In this work, we introduce GEM-x, a Gaussian-mixture flow-matching framework for crystal structure generation. GEM-x operates in an atom-wise latent representation space \citep{joshi2025allatom} that enhances structural diversity while maintaining energetic plausibility. In the latent space, each latent vector encodes information needed to reconstruct the composition, lattice and atomic arrangement of a crystal. GEM-x approximates the latent denoising distribution as a Gaussian mixture. This multi-modal parameterization allows the model to represent multiple plausible denoising trajectories from the same noisy state and, therefore, explore composition–structure space more broadly without sacrificing energetic plausibility. GEM-x is implemented with sparsely gated mixture-of-experts layers \citep{shazeer2017outrageously}, where expert routing is conditioned on the denoising noise level and target space group, enabling different experts to specialize across denoising regimes and crystal symmetries. 
Trained on the MP-20 dataset, GEM-x generates stable or metastable, unique, and novel structures at a rate of 12.3\%, achieving a 46\% improvement in the generation of metastable, unique, and novel structures over latent flow-matching models. All results are validated using $r^2$SCAN density functional theory (DFT) calculations, which provide substantially improved formation-energy accuracy relative to conventional GGA functionals \cite{kothakonda2023r2scan}. The Results section provides an overview of the GEM-x framework, evaluates its performance against existing generative models, and presents ablation studies of its key components. The Discussion section examines the implications of these findings for materials discovery, while the Methods section details the model formulation and implementation.

\begin{figure}[htbp]
    \centering
    \includegraphics[width=1.2\textwidth]{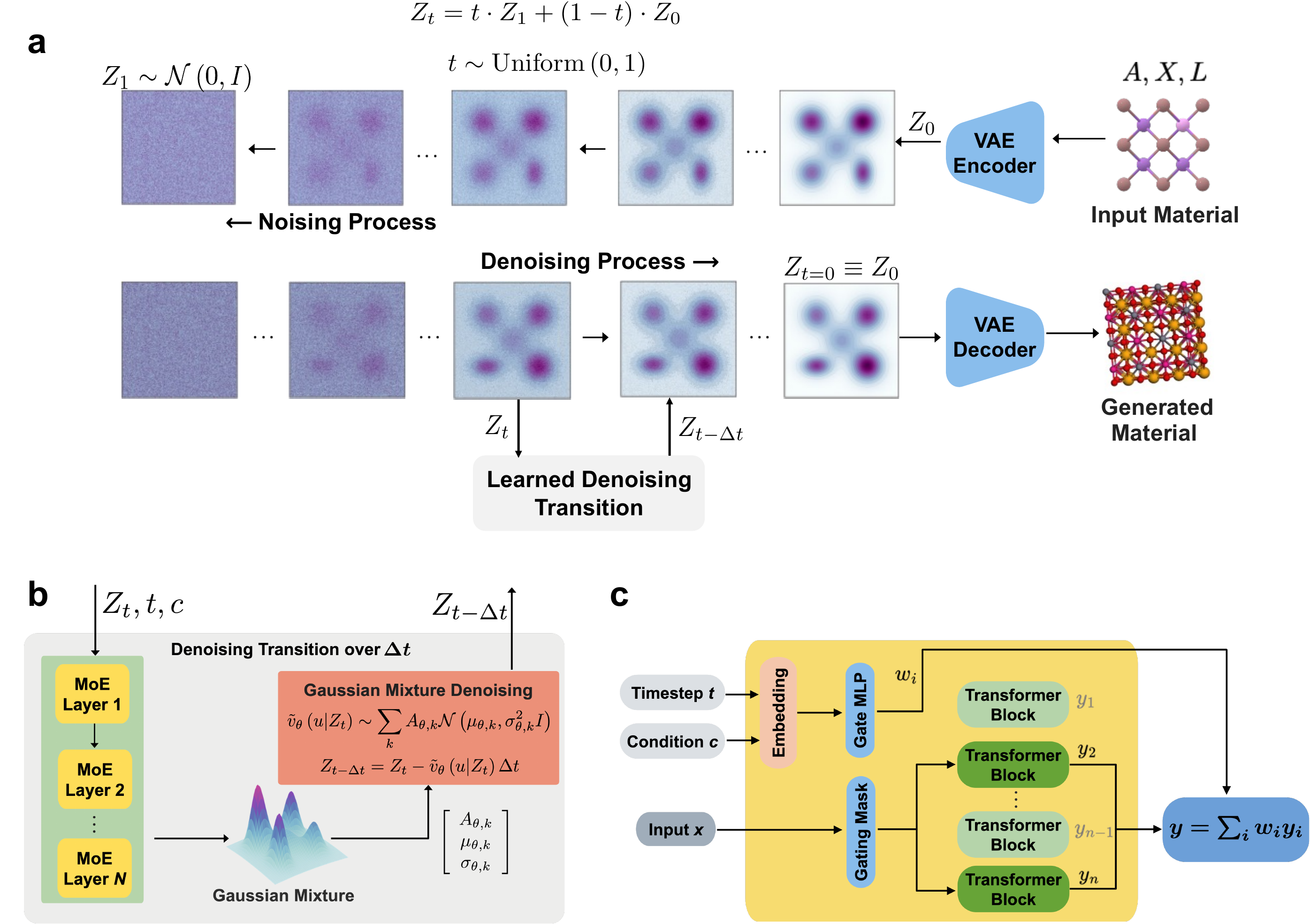}

    \caption[Exploration-driven Materials Discovery using Gaussian Mixtures and Ensemble Learning]{%
        \textbf{Exploration-driven Materials Discovery using Gaussian Mixtures and Ensemble Learning.}
        \textbf{(a)} A crystal structure, represented by atom types $A$, atomic positions $X$, and lattice tensor $L$, is encoded into a latent representation $Z_0$ using the pre-trained variational autoencoder of the all-atom diffusion transformer (ADiT) framework \citep{joshi2025allatom}. During noising, $Z_0$ is interpolated with Gaussian noise $Z_1$ to obtain intermediate latent states $Z_t$. GEM-x learns the denoising process, which, upon decoding the latent representation at $t=0$, yields a new generated material. \textbf{(b)} GEM-x learns the reverse denoising process using successive Mixture-of-Experts (MoE) layers. The resulting network predicts a Gaussian mixture parameterized by weights $A_{\theta,k}$, means $\mu_{\theta,k}$, and standard deviations $\sigma_{\theta,k}$. A denoising velocity $\tilde{v}_\theta$ sampled from this distribution updates $Z_t$ to $Z_{t-\Delta t}$. \textbf{(c)} Each MoE layer comprises Transformer-based experts and a learnable gating mechanism. The gating network uses timestep $t$, conditioning information $c$, and input features $x$ to route information through selected experts, whose outputs $y_i$ are combined as $y=\sum_i w_i y_i$.
    }
    \label{fig:arch_complete}
\end{figure}
\clearpage
% \end{landscape}

\section*{Results}

\textbf{Motivation and framework.} Effective materials discovery requires the generation of novel, energetically plausible crystal structures, encompassing both thermodynamically stable and potentially accessible metastable phases. Importantly, novelty requires exploring regions of the composition–structure space beyond those represented in the training data. However, as discussed earlier, flow-based generative models may preferentially reproduce known structures or generate candidates with substantial similarity to their training examples, limiting their ability to discover genuinely new materials. Although such structures may be energetically plausible, repeatedly generating known structures offers limited value for materials discovery.  Additionally, a higher proportion of duplicate structures reduces sampling efficiency, requiring more model evaluations to generate the same number of unique and novel candidates. Consequently, generative materials discovery requires balancing energetic plausibility with structural novelty to enable the identification of previously unexplored stable and metastable crystalline materials.

We address this challenge with GEM-x, a Gaussian mixture flow matching framework designed to achieve the aforementioned balance.  The framework is depicted in Figure \ref{fig:arch_complete}. GEM-x parametrizes the distribution over denoising flow velocities with a mixture of $k$ Gaussians in the latent space, which allows for multiple stochastic trajectories from the same latent state. This enables greater exploration of the composition-structure landscape and denoising to hitherto unseen regions of the data manifold. As seen in Fig. \ref{fig:arch_complete} ,  sparsely gated mixture-of-experts layers \cite{shazeer2017outrageously} are used to parametrize the mean ($\mu_{\theta, k}$), variance ($\sigma_{\theta, k}$), and mixture weights ($A_{\theta, k}$) of the Gaussian Mixture. Each layer in turn consists of $M$ experts, which are transformer modules. A gating network selects a subset of experts from each layer, for each noised latent state. The gating mechanism conditions the expert routing on the denoising time and the target spacegroup, enabling different experts to specialize across noise levels and symmetry-conditions. This architecture significantly increases the total expressive capacity of the model without the proportional increase in computational cost.

\noindent \textbf{Dataset, performance metrics, and benchmarks.} To train GEM-x, we use the MP-20 dataset, which represents 45{,}231  crystals from the Materials Project \citep{jain2013}, with up to 20 atoms in a unit cell containing 89 different elements. These materials differ in both structure and composition. Furthermore, all crystals in this dataset have been relaxed using DFT calculations and are either stable or metastable, which is critical for generation of energetically plausible crystals.  Training details have been provided in Supplementary Note 1.

 We evaluate GEM-x using metrics that measure physical feasibility, energetic plausibility, and diversity of generated samples. Namely, we evaluate validity, uniqueness, novelty, and stability of generated crystal structures. Validity measures the fraction of generated structures satisfying basic geometric and chemical constraints. This metric is a primary screening criterion and is indicative of adherence to the manifold of materials. A crystal is said to be valid if it is structurally and compositionally valid. Structural validity is assessed by checking whether all pairwise interatomic distances are at least $0.5\,\text{\AA}$ and the unit cell volume is at least $0.1\,\text{\AA}^3$. Compositional validity is determined by checks on whether a crystal is charge neutral, element valencies are satisfied, and electro-negativity rules are satisfied \citep{joshi2025allatom}. Uniqueness measures the percentage of generated structures that are distinct from one another. Novelty measures the percentage of structures that are absent from the training set. Uniqueness and novelty are used to quantify exploration beyond the data distribution.  
 
 A generated crystal structure is considered energetically plausible if it is either thermodynamically stable, or lies within a specified metastability window. Thermodynamic stability is assessed using the energy above the convex hull, $E_{\mathrm{hull}}$, which measures the energy difference between a candidate structure and the lowest-energy competing phase assemblage at the same composition \cite{jain2013}. Candidate structures are relaxed using DFT calculations, and their energies are subsequently evaluated relative to the thermodynamic convex hull. In this work, DFT calculations are performed using the $r^2$SCAN meta-GGA \cite{furness2020r2scan} functional, which has demonstrated substantially improved formation-energy accuracy relative to conventional PBE-GGA \cite{perdew1996pbe} calculations, while providing comparable or improved accuracy for decomposition energies depending on the chemical system \citep{kothakonda2023r2scan}. A structure is classified as thermodynamically stable if its DFT-calculated energy lies on or below the reference convex hull ($E_{\mathrm{hull}} \leq 0$). In contrast, metastable structures have positive energies above the convex hull, but may remain experimentally accessible due to kinetic barriers \citep{sun2016metastability,aykol2018thermodynamic}.  Although the energetic range of experimentally accessible metastable structures varies considerably across chemical systems, an energy-above-hull threshold of 0.1 eV/atom is commonly employed as a heuristic screening criterion for identifying potentially accessible metastable candidates, rather than as a definitive measure of synthesizability \citep{bartel2018gibbs,tolborg2022freeenergy,twyman2022environmental,chen2022universal,Zeni2025MatterGen}. Accordingly, we define metastable structures as those satisfying $0 < E_{\mathrm{hull}} \leq 0.1$ eV/atom. We evaluate generative performance using two metrics: the stable, unique, and novel (SUN) rate and the metastable, unique, and novel (MSUN) rate. SUN denotes the fraction of generated structures that are unique, novel, and thermodynamically stable ($E_{\mathrm{hull}} \leq 0$), whereas MSUN denotes the fraction of structures that are unique, novel, and metastable ($0 < E_{\mathrm{hull}} \leq 0.1$ eV/atom). This distinction ensures that the two metrics independently quantify the generation of thermodynamic ground-state candidates and potentially accessible metastable structures.

We compare the structural and compositional validity rates of GEM-x against crystal generative models, including CDVAE \citep{xie2022}, DiffCSP \citep{jiao2023diffcsp}, UniMat \citep{yang2024}, FlowMM \citep{miller2024}, FlowLLM \citep{sriram2024}, and ADiT \citep{joshi2025allatom}. We further present a detailed comparison with ADiT, a leading benchmark for crystal structure generation on MP-20, evaluating structural and compositional validity, valid, unique, and novel (VUN) rates, and energy-above-hull distributions. Since thermodynamic stability evaluations have commonly relied on PBE/PBE+U energies, direct comparison with GEM-x, which employs the $r^2$SCAN meta-GGA functional, is not straightforward. We therefore independently reproduce ADiT's DFT evaluation using the same $r^2$SCAN workflow employed for GEM-x, enabling a consistent comparison of the thermodynamic stability and metastability of generated structures. This comprehensive evaluation allows us to assess not only the physical plausibility of generated structures but also their diversity, novelty, and energetic plausibility relative to a leading generative baseline.

\begin{figure}[t!]
    \centering    \includegraphics[width=0.8\linewidth]{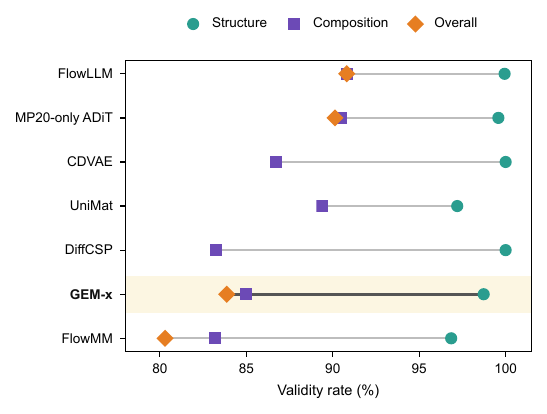}
    
    \caption{Comparison of structural, and compositional validity across crystal generation models. The percentage of structures being both structurally and compositionally valid is labeled as overall.  Each row corresponds to a model, and markers indicate the reported validity rates. If structural validity is 100\%, then overall validity is simply the compositional validity. Overall validity is not the product of structural and compositional validity, as these need not be independently distributed. Missing values correspond to metrics not reported in the source literature.}
    
    \label{fig:validity-structure-composition-overall}
\end{figure}

\noindent \textbf{Model performance analysis.} Figure \ref{fig:validity-structure-composition-overall} compares the structural and compositional validity rates of GEM-x with those of the benchmark models. Since all models take atomic positions and lattice tensors as inputs and build their internal representations around these structural features, structural validity rates are consistently near 100\%. In contrast, compositional validity rates are substantially lower, as most generative models do not explicitly account for atomic valencies, electronegativity, or oxidation states. Instead, these chemical constraints must be learned from the training data distribution. Consequently, compositional validity provides a measure of how well
the generated structures adhere to the underlying crystal manifold.

\begin{figure}[t!]
    \centering
    \includegraphics[
        width=0.8\linewidth,
    ]{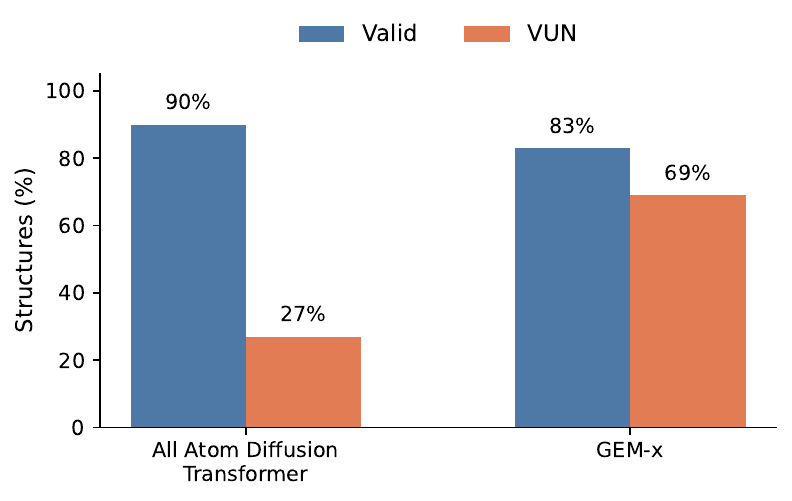}
    \caption{
        Comparison of validity and VUN rates for ADiT and GEM-x.
        Bars indicate the percentage of generated structures satisfying each criterion.
    }
    \label{fig:valid_vun_comparison}
\end{figure}

While GEM-x has a lower validity rate, as shown in
Figure~\ref{fig:valid_vun_comparison}, higher validity rates
do not necessarily translate to more effective exploration of
crystal space. Although ADiT achieves a higher validity rate than GEM-x (90 vs 83\%), its substantially lower VUN rate (27\% versus 69\%) highlights an important trade-off between validity and diversity of generated crystals. In particular, improvements in validity may yield diminishing returns if they are accompanied by increased duplication or reproduction of structures from the training distribution. When a model repeatedly generates the same or previously known structures, a larger number of samples must be generated to obtain a given number of unique and novel candidates, reducing the efficiency of the generation process for materials discovery. Conversely, greater exploration of crystal space can increase the likelihood of generating structures that deviate from the underlying crystal manifold. Thus, maximizing validity in isolation is not necessarily more conducive to the goals of materials discovery; rather, validity must be balanced with uniqueness and novelty to enable efficient discovery of diverse, physically plausible crystal structures.

\begin{figure}[t!]
    \centering

    \begin{subfigure}[t]{0.8\textwidth}
        \centering
        \includegraphics[
            width=\linewidth
        ]{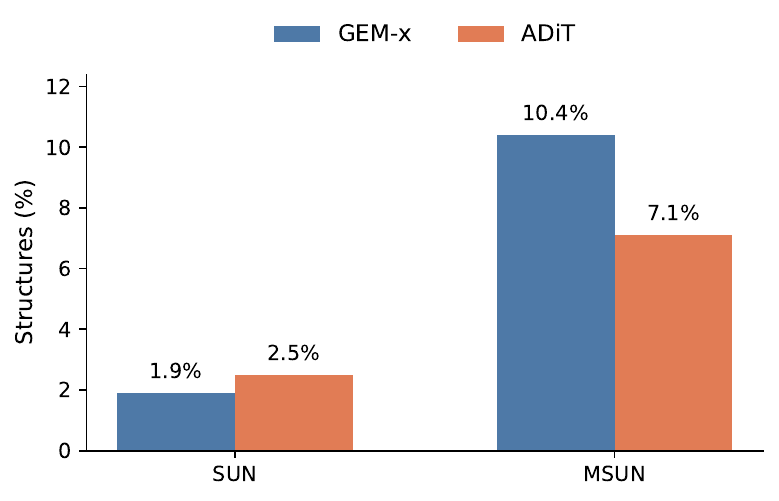}
        \caption{SUN and MSUN rates.}
        \label{fig:rate-comparisons}
    \end{subfigure}
    \hfill
    \begin{subfigure}[t]{0.8\textwidth}
        \centering
        \includegraphics[
            width=\linewidth
        ]{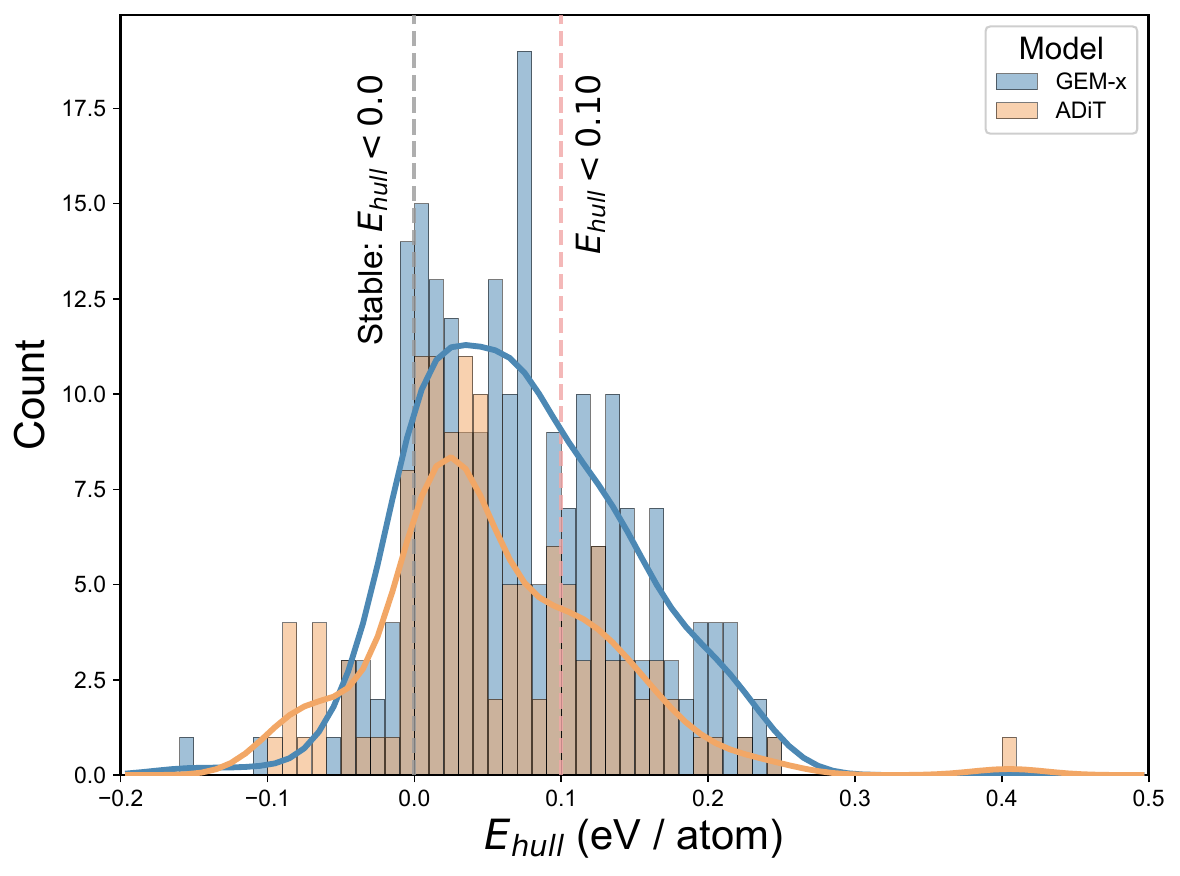}
        \caption{Energy-above-hull distributions of generated structures.}
        \label{fig:hull_distribution}
    \end{subfigure}

    \caption{
        Thermodynamic stability of crystal structures generated
        by GEM-x and ADiT, evaluated using $r^2$SCAN DFT.
        (a) Percentage of generated structures classified as
        stable, unique, and novel (SUN) or metastable, unique,
        and novel (MSUN).
        (b)Histogram for distribution of energies above the convex hull, along with the  best fit curve based on Gaussian smearing for both ADiT and GEM-x.
    }
    \label{fig:hull}

\end{figure}

Thermodynamic stability is another key criterion for evaluating the performance of crystal structure generative models. As shown in Figure \ref{fig:rate-comparisons}, GEM-x achieves a SUN rate of 1.9\%, compared with 2.5\% for ADiT. However, GEM-x substantially outperforms ADiT in the generation of MSUN structures, achieving a rate of 10.4\% compared with 7.1\% for ADiT, representing a 46\% improvement. Given the importance of metastable crystalline phases in materials discovery, evaluating performance solely in terms of the generation of thermodynamic ground states provides an incomplete picture of a model's ability to discover energetically plausible materials. We therefore consider the combined SUN and MSUN rate, which captures the generation of both thermodynamically stable and potentially accessible metastable structures. Under this metric, GEM-x achieves a rate of 12.3\%, compared with 9.6\% for ADiT, corresponding to a 28\% improvement. These results demonstrate that, although GEM-x produces slightly fewer thermodynamic ground-state candidates, it substantially increases the overall yield of novel, energetically plausible crystal structures, highlighting its effectiveness in exploring the broader landscape of stable and metastable materials.

\begin{figure}[t]
    \centering

    % Element distribution
    \begin{subfigure}[t]{0.8\textwidth}
        \centering
        \includegraphics[
            width=\linewidth,
            height=5.5cm,
            keepaspectratio
        ]{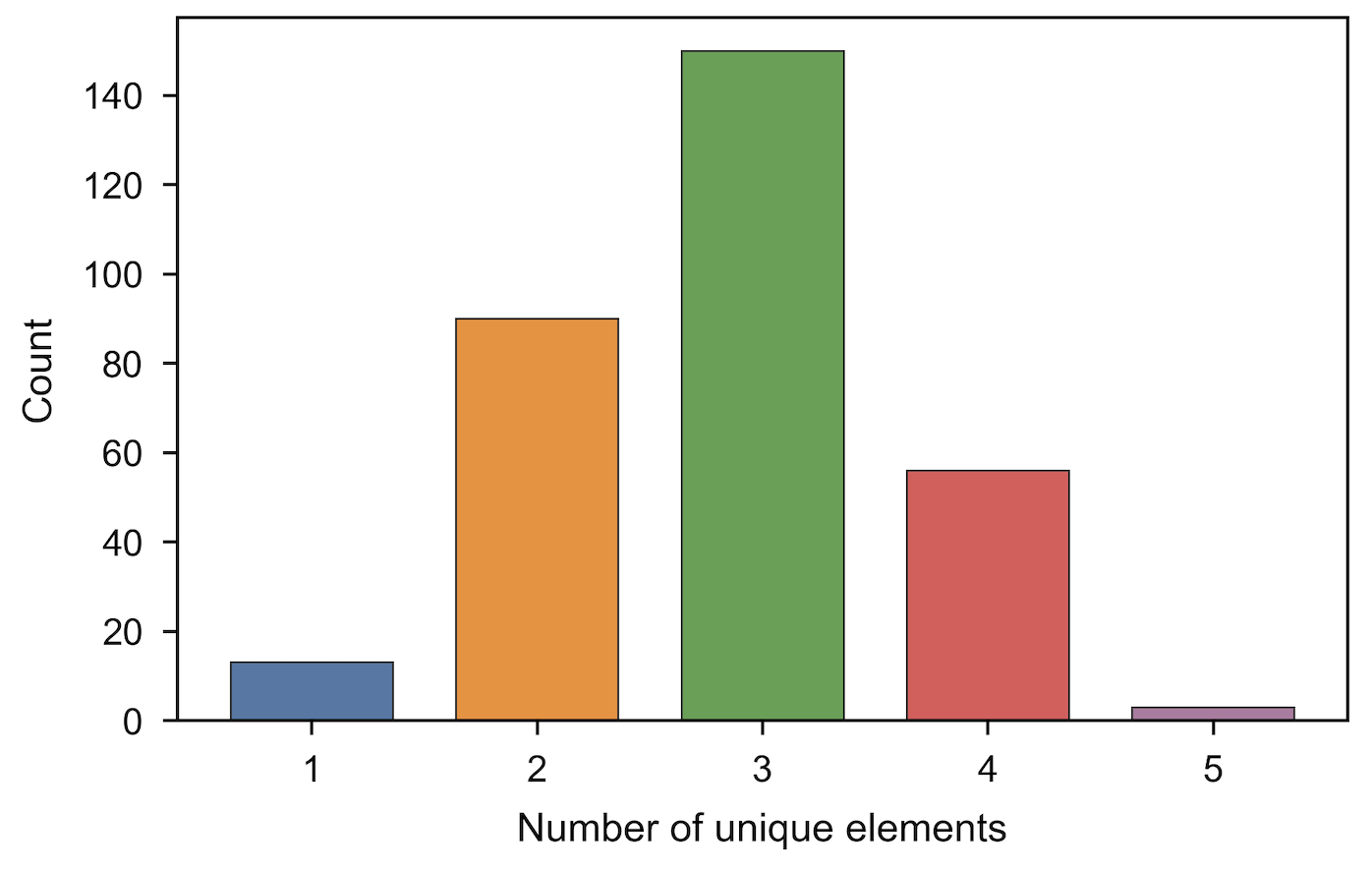}
        \caption{Distribution of unique elements per crystal.}
        
        \label{fig:element_distribution}
    \end{subfigure}

    \vspace{1em}

    % Representative crystal structures
    \begin{subfigure}[t]{0.96\textwidth}
        \centering

        \begin{minipage}[c]{0.30\linewidth}
            \centering
            \includegraphics[
                width=\linewidth,
                height=3.8cm,
                keepaspectratio
            ]{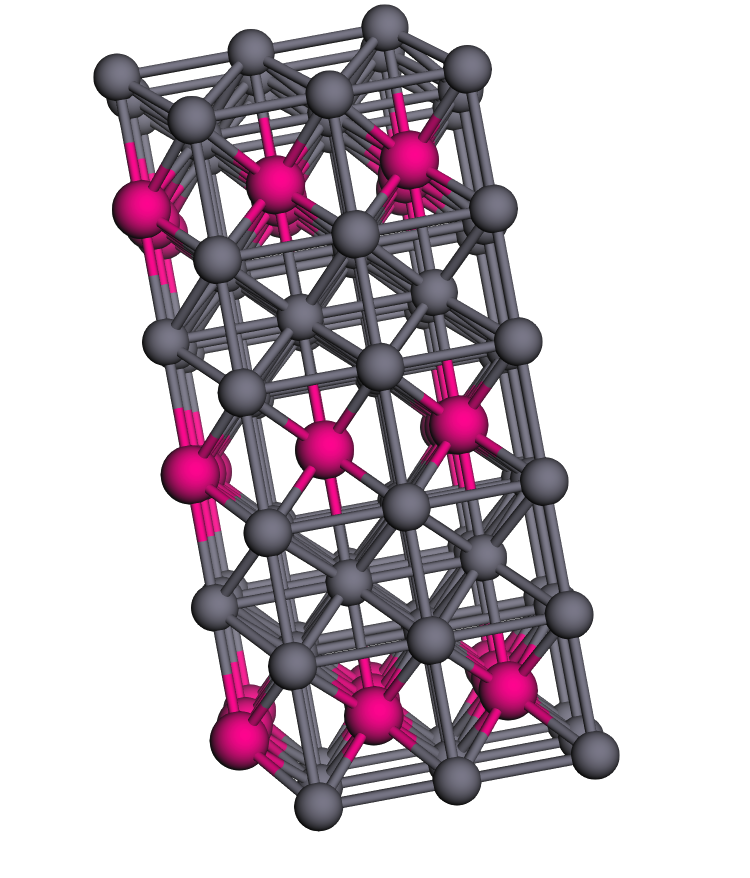}

            \vspace{0.3em}
            $\mathrm{Cr_3Ga}$
        \end{minipage}
        \hfill
        \begin{minipage}[c]{0.30\linewidth}
            \centering
            \includegraphics[
                width=\linewidth,
                height=3.8cm,
                keepaspectratio
            ]{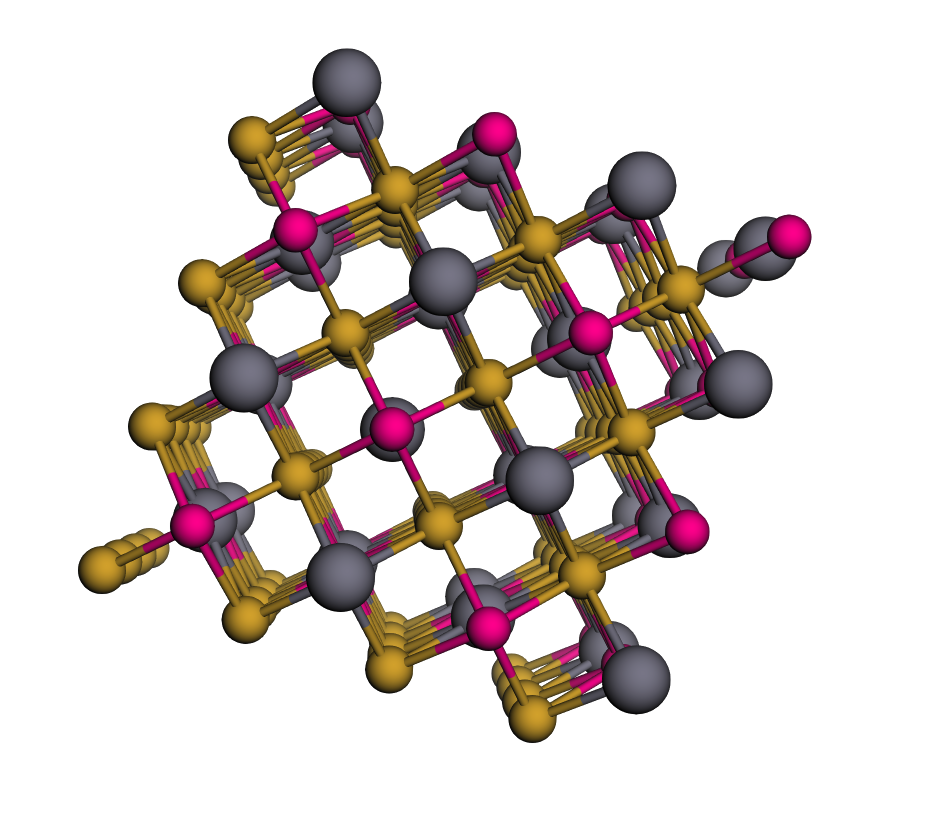}

            \vspace{0.3em}
            $\mathrm{CaEuAu_2}$
        \end{minipage}
        \hfill
        \begin{minipage}[c]{0.30\linewidth}
            \centering
            \includegraphics[
                width=\linewidth,
                height=3.8cm,
                keepaspectratio
            ]{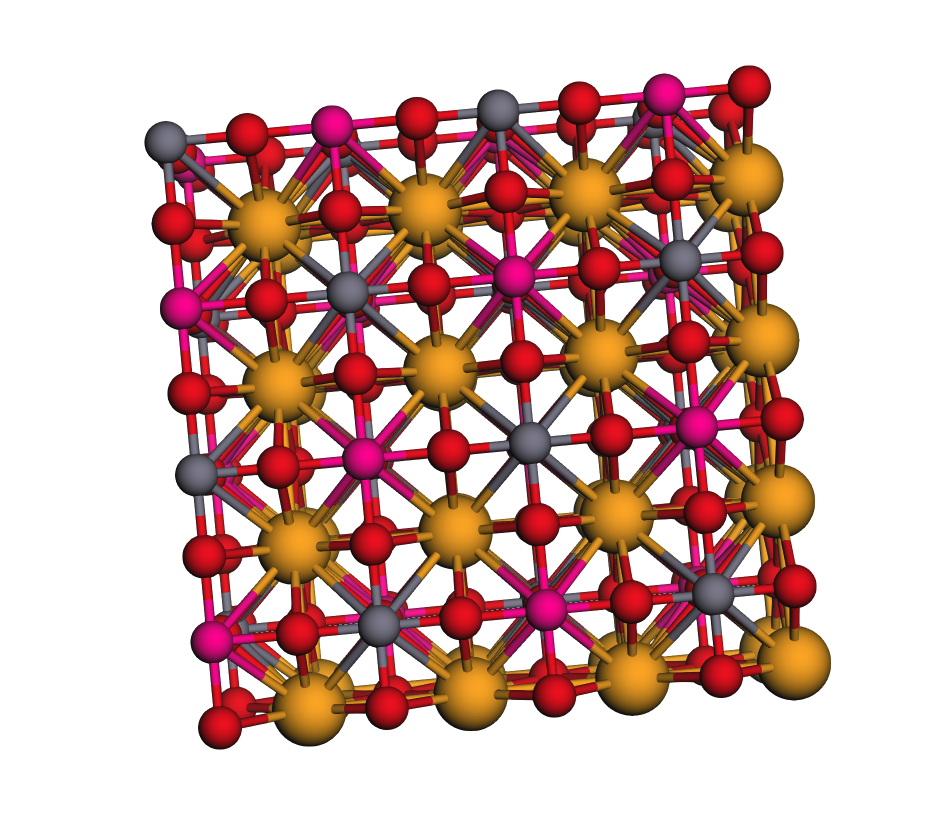}

            \vspace{0.3em}
            $\mathrm{Ba_2HfTiO_6}$
        \end{minipage}

        \caption{Representative crystal structures generated by GEM-x.}
        \label{fig:generated_structures}
    \end{subfigure}
\caption{
    Compositional diversity and representative crystal structures
    generated by GEM-x.
    (a) Distribution of the number of distinct elements per
    generated crystal.
    (b) Representative crystal structures generated by GEM-x,
spanning binary, ternary, and quaternary chemical systems.
The binary compound $\mathrm{Cr_3Ga}$ is thermodynamically stable,
whereas the ternary $\mathrm{CaEuAu_2}$ and quaternary
$\mathrm{Ba_2HfTiO_6}$ compounds are metastable, as determined
by $r^2$SCAN DFT calculations.
}

\end{figure}
Figure \ref{fig:hull_distribution} compares the distributions of energy above the convex hull for GEM-x and ADiT, evaluated using $r^2$SCAN DFT. Both models initially generated 1,000 candidate structures, which underwent filtering and pre-relaxation before DFT evaluation, as detailed in Supplementary Note 2. Since invalid and duplicate structures were removed during prefiltering, the distributions represent only the candidates retained for DFT calculations. The histograms display raw counts rather than normalized probability densities, so the area within a given energy interval is proportional to the number of structures in that interval. GEM-x exhibits a substantially larger area in the metastable region ($0 < E_{\mathrm{hull}} \leq 0.1$ eV/atom), reflecting a greater number of energetically plausible candidates retained from the same initial number of generated structures.

The mean and median hull energies for GEM-x are 0.0706 and 0.0670 eV/atom, respectively, compared with 0.0523 and 0.0390 eV/atom for ADiT. The standard deviations are comparable: 0.0747 eV/atom for GEM-x and 0.0766 eV/atom for ADiT. ADiT's distribution is more strongly concentrated near the convex hull, indicating a greater tendency for its retained candidates to exhibit lower hull energies. However, as discussed above, the lower frequency of duplicate structures generated by GEM-x allows a larger number of distinct candidates to pass through prefiltering and proceed to DFT evaluation. This higher throughput of retained structures contributes to the greater absolute number of metastable candidates observed in Figure \ref{fig:hull_distribution}.  Importantly, this shows that generating structures closer to the convex hull does not necessarily result in more energetically plausible candidates overall, particularly when a substantial fraction of generated structures are duplicates. By prioritizing exploration while maintaining a substantial fraction of energetically plausible structures, GEM-x converts a greater proportion of its initial generations into distinct candidates with potential relevance for materials discovery. This further emphasizes the importance of balancing exploration of the composition–structure space with adherence to the underlying manifold in materials discovery.

Furthermore, the distribution of the number of distinct elements per generated crystal, shown in Figure \ref{fig:element_distribution}, demonstrates that GEM-x generates structures across a broad range of compositional complexities, rather than disproportionately favoring unary systems or compounds containing many distinct elements. To further illustrate this compositional diversity, Figure \ref{fig:generated_structures} presents three representative crystals generated by GEM-x: $\mathrm{Cr_3Ga}$, $\mathrm{CaEuAu_2}$, and $\mathrm{Ba_2HfTiO_6}$, spanning binary, ternary, and quaternary chemical systems, respectively. These examples provide a structural illustration of GEM-x's ability to generate crystals with varying numbers of constituent elements.

\noindent \textbf{Ablation studies.} We investigated the key components of GEM-x, namely,  the number of Gaussians in the denoising distribution, the stochasticity in the denoising dynamics, and the sparsity of the MoE. The number of Gaussians in the denoising distribution governs the extent of multi-modality that the model captures in the denoising process. Ablating for the stochasticity in the denoising dynamics allows us to ascertain if a stochastic denoising trajectory is more effective for the exploration of the composition-structure landscape, while balancing adherence to the learned manifold. Varying the sparsity of the MoE illustrates whether the increased expressive power of ensemble learning is more advantageous in approximating the multi-modal denoising distribution.

\begin{figure}[t!]
    \centering
    \includegraphics[width=0.8\linewidth]{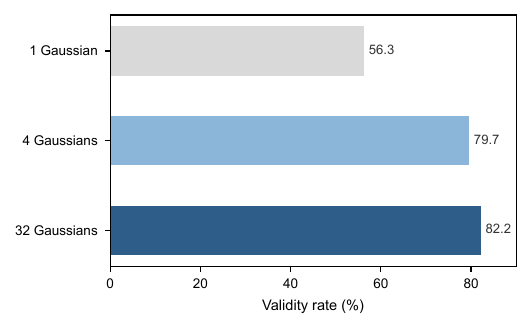}
    \caption{
    Effect of the number of Gaussian mixture components on validity rate.
    The tested configurations contain 1, 4, and 32 Gaussian components. It is evident that beyond 4 Gaussians, additional modes only lead to diminishing returns.
    }
    \label{fig:gaussian-ablation-validity}
\end{figure}

\emph{Number of components in the Gaussian Mixture.} To verify the multi-modal nature of the denoising distribution, we varied the number of Gaussian components predicted by the model. Here, the key metric to evaluate is the validity rate. This is because underestimation of the number of modes in the denoising distribution would lead to crude approximations of the data and, therefore, produce a greater fraction of off-manifold structures. Figure \ref{fig:gaussian-ablation-validity} shows the validity rates for models with 1, 4 and 32 Gaussian modes in their denoising distribution.  Validity rates increase from 56.3\% to 82.2\% with an increase in the number of components in the Gaussian mixture, from 1 to 32. A single Gaussian is equivalent to denoising-diffusion probabilistic modeling (DDPM) \cite{ho2020denoising} in the latent space. The change in validity rates from 4 Gaussians to 32 indicates that overestimating the number of modes in the denoising distribution provides diminishing returns in manifold approximation. These results highlight the importance of multimodality in crystal structure generation, suggesting that the ability to represent multiple plausible generative trajectories from a shared latent state enhances the model's capacity to generate physically plausible crystal structures.

\begin{figure}[t!]
    \centering

    \begin{subfigure}[t]{0.8\linewidth}
        \centering
        \includegraphics[width=\linewidth]{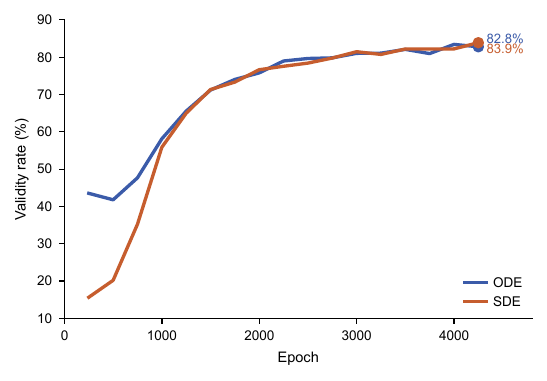}
        \caption{Validity rates vs epochs during training}
        \label{fig:ode-sde-training-validity}
    \end{subfigure}
    \hfill
    \begin{subfigure}[t]{0.8\linewidth}
        \centering
        \includegraphics[width=\linewidth]{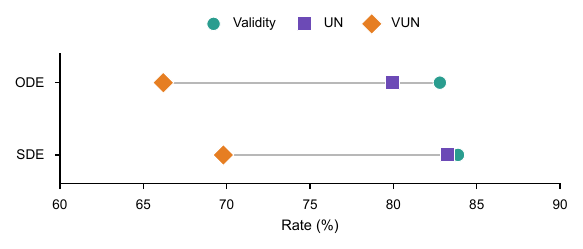}
        \caption{Terminal sample quality.}
        \label{fig:ode-sde-terminal-vun}
    \end{subfigure}

    \caption{
    Comparison of deterministic ODE and stochastic SDE sampling.
    \textbf{a}. Validity rate during training for ODE and SDE sampling, with terminal values taken at epoch 4250.
    \textbf{b}. Terminal validity, UN, and VUN rates for ODE and SDE sampling.
    UN is computed as VUN divided by the terminal validity rate.
    }
    \label{fig:ode-sde-deterministic-comparison}
\end{figure}

\emph{Stochasticity in denoising dynamics.}
The nature of denoising dynamics has a significant impact on the quality and diversity of generated crystal structures. We considered two models: one in which denoising trajectories were obtained by solving an ordinary differential equation (ODE) of flow velocities obtained by taking an expectation over all Gaussian modes, and a stochastic denoising model where trajectories were obtained by sampling from the Gaussian mixture at every step. We compare the validity rates for both models against the number of training epochs in Figure \ref{fig:ode-sde-training-validity}. In the ODE model, the validity rates increased significantly earlier in training. This is because the ODE model has a greater manifold adherence since, after the initial sampling from the standard Gaussian, the denoising trajectory follows a deterministic flow path. In the stochastic model, the randomness in the denoising dynamics causes sampling of some structures that lie off manifold, leading to lower validity rates in initial epochs of training.  Upon further training, the difference in sample quality between the two cases vanishes due to the alignment of Gaussian modes with true flow velocities, which biases sampling in favor of producing valid structures as substantiated in Figure \ref{fig:ode-sde-training-validity}. Figure \ref{fig:ode-sde-terminal-vun} shows that a notable difference exists in the diversity of generated structures in both cases. The stochastic case shows a higher uniqueness and novelty rate, exceeding the deterministic case by nearly 5\%. The negligible difference in validity and the superior sample diversity make the stochastic denoising model better suited for materials discovery.

\clearpage
\begin{figure}[H]
    \centering
    \includegraphics[width=0.8\linewidth]{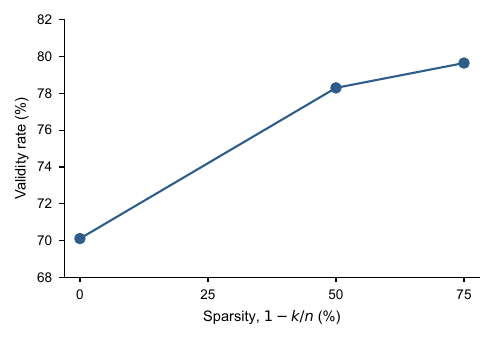}
    \caption{
    Effect of sparse mixture-of-experts (MoE) routing on validity rate.
    Sparsity is defined as \(1-k/n\), where \(n\) is the total number of experts and \(k\) is the number of selected experts.
    The three configurations correspond to 2 experts with both selected, 4 experts with top-2 routing, and 8 experts with top-2 routing. For all models, the number of gradient updates per expert are held constant.
    }
    \label{fig:sparsity-ablation-validity}
\end{figure}

\emph{MoE sparsity.} To test the expressive power of the ensemble of experts, we ablated for the total number of callable experts while maintaining the number of selected experts constant. This comparison ensured that the total number of active parameters is the same across all models. We define sparsity as $(1-k/n)$, where $k$ is the number of selected experts and $n$ is the total number of callable experts, and ablate over $n$ while keeping $k=2$. It is important to note that, in a denser model, individual experts receive gradient updates more frequently than in a sparser one. Thus, we have trained the sparser models for a proportionately higher number of epochs. Figure \ref{fig:sparsity-ablation-validity} compares the validity rates across three different sparsity values, with validity rate increasing with sparsity.  The validity rates improve significantly for sparser models with higher number of total parameters. This is because the total number of callable parameters for these models is significantly higher, even if the number of active parameters remains the same. For example, from the dense case to the sparsest case, the total number of callable parameters increases from 255M to 1.1B. The gating mechanism of expert selection also allows different experts to specialize at different stages in the denoising trajectory and at different conditions provided for generation. Furthermore, expert activation heatmaps, shown in Supplementary Note 3, reveal that experts specialize based on spacegroup conditioning and the timestep in the trajectory. The higher expressive power of the ensemble enables for better manifold approximation owing to the increased number of callable parameters in the model.

\section*{Discussion}

Generative models offer a promising approach to materials discovery. However, realizing the potential of these models requires balancing adherence to the manifold of physically plausible crystal structures with the capacity to explore novel configurations beyond those represented in the training data. While adherence to the underlying manifold is essential for generating physically plausible structures, it does not necessarily translate into a greater number of useful candidates for materials discovery. As illustrated by Figure \ref{fig:valid_vun_comparison}, higher validity can coexist with substantial duplication or reproduction of previously known structures, limiting the number of novel candidates obtained from a given number of generations. Similarly, evaluating thermodynamic stability at 0 K alone may not fully capture the potential of generated materials. Metastable phases can be experimentally accessible and may exhibit desirable properties that are absent in their thermodynamically stable counterparts. Moreover, a structure that is energetically metastable at 0 K may become thermodynamically stable at finite temperatures, as entropic contributions can alter the relative stability of competing phases \citep{bartel2018gibbs}. For example, $\mathrm{BaTiO_3}$ undergoes temperature-dependent structural phase transitions, with the rhombohedral phase stable at low temperatures and the tetragonal phase stable at room temperature \citep{gigli2022batio3}. Our results further demonstrate how including metastable candidates provides a more comprehensive assessment of generative performance. While ADiT achieves a  higher SUN rate, GEM-x generates a higher proportion of MSUN structures, resulting in a higher combined rate of energetically plausible structures. Thus, evaluating only thermodynamically stable structures would overlook a substantial fraction of the unique, novel, and energetically plausible candidates generated by GEM-x. These considerations highlight the importance of balancing exploration with manifold adherence and evaluating generative models by their overall yield of unique, novel, and energetically plausible crystal structures, rather than focusing exclusively on validity or thermodynamic ground-state stability.

GEM-x addresses this balance between exploration and manifold adherence. Firstly, the parametrization of denoising flow velocities using a Gaussian mixture captures multiple modes of the denoising distribution, leading to multiple plausible crystal structures from the same noised latent state. Secondly, sampling from a Gaussian mixture introduces stochasticity in the denoising process, which yields a greater diversity of generated structures. Together, these two factors promote generation in underrepresented regions of the composition–structure space, reducing the frequency of duplicate structures. This increased exploration, however, introduces a trade-off: while it enhances structural diversity and the likelihood of discovering novel candidates, it may also increase the proportion of off-manifold and thermodynamically unstable structures. Nevertheless, the observed increase in unique, novel, and energetically plausible structures suggests that the benefits of broader exploration can outweigh this potential reduction in stability, ultimately yielding more distinct candidates for materials discovery. Additionally, the sparsely gated MoE architecture enables GEM-x to leverage conditional computation, wherein the choice of expert selection and weight is guided by the noise level in a given latent state and the conditioning provided upon generation. A consequence of this conditional computation is expert specialization based on the guidance provided for expert selection.

The thermodynamic assessment of the generated candidates also warrants careful consideration. In this work, we evaluate thermodynamic stability using $r^2$SCAN DFT, which has demonstrated improved formation-energy accuracy relative to conventional PBE-GGA methods employed in many previous benchmarks \citep{kothakonda2023r2scan}. The use of $r^{2}$SCAN DFT provides a more reliable assessment of the energetic plausibility of generated structures. Nevertheless, the present DFT calculations evaluate stability at 0 K and do not account for finite-temperature contributions to the Gibbs free energy, kinetic effects, or experimental synthesis conditions. Consequently, the calculated energy above the convex hull cannot independently establish whether a generated structure is experimentally synthesizable, or thermodynamically stable at room temperature. Furthermore, the 0.1 eV/atom metastability threshold is a heuristic screening criterion, rather than a universal boundary for synthesizability, as the energetic accessibility of metastable phases varies across chemical systems \citep{aykol2018thermodynamic}. The structures identified in this work should therefore be regarded as promising candidates whose experimental accessibility requires further investigation.

An important direction for future work is the development of more effective conditional alignment methods for targeted materials discovery. While the present framework focuses on generating diverse, energetically plausible crystal structures, subsequent fine-tuning could enable generation toward specific material properties or combinations of properties. Reinforcement learning offers a promising approach in this regard, allowing pretrained generative models to be optimized using reward signals derived from predicted material properties and thermodynamic stability \citep{cao2026reinforcement,huang2026formation,chen2025matinvent}. Recent studies have demonstrated the feasibility of reinforcement learning-based fine-tuning for property-guided crystal generation, including objectives involving thermodynamic stability and desired functional properties. As machine-learned interatomic potentials (MLIPs) become more accurate and computationally efficient, they could provide increasingly reliable estimates of structural energies and stability to guide generative models. Specifically, MLIP-based energy evaluations could be incorporated into the fine-tuning process as reward signals, encouraging the generation of structures with favorable energetic properties. Integrating such feedback into the generative process could facilitate more targeted exploration of the composition–structure space and accelerate the identification of promising materials for further investigation.

\section*{Methods}

GEM-x is a latent flow-based generative framework that combines Gaussian Mixture Flow Matching \citep{chen2025gaussian} with a sparsely gated Mixture-of-Experts (MoE) architecture \citep{shazeer2017outrageously}. As illustrated in Figure~\ref{fig:arch_complete}, crystal structures are first encoded into a continuous latent space, where the generative process is formulated as a flow from a Gaussian prior $Z_1$ toward the data distribution $Z_0$. Rather than approximating a single flow velocity, GEM-x learns a conditional distribution $q_\theta(u \mid Z_t,t,c)$ over denoising velocities, parameterized as a Gaussian mixture. The distribution is predicted by successive MoE layers comprising Transformer-based experts, with sparse expert selection conditioned on the timestep $t$ and target space group $c$. During inference, velocities sampled from the learned distribution guide the reverse flow, producing latent representations that are decoded into crystal structures. The following sections detail the latent representation, flow-matching formulation, training objective, conditional guidance, denoising dynamics, and network architecture.

\par\medskip
\noindent\textbf{Latent space construction and flow matching priors.}
A material structure is represented by atomic species $A$, Cartesian coordinates $X$, fractional coordinates $F$, and lattice vectors $L$. Following the unified atom-wise representation of ADiT \citep{joshi2025allatom}, we use a pretrained variational autoencoder (VAE) to map this combined discrete--continuous representation to a continuous latent space. The encoder produces a set of per-atom latent vectors
\[
Z_{0} = E(A,X,F,L), \qquad 
Z_{0}=\{z_i^{(0)}\}_{i=1}^{N}, \quad z_i^{(0)}\in\mathbb{R}^{d}.
\]
Where $N$ represents the total number of atoms in a crystal structure, and $d$ is the latent dimension of the VAE. This representation enables continuous, differentiable diffusion trajectories while adhering to the inductive bias that an atom-wise representation captures contextual information from different atoms in the same material. For further details, refer to Supplementary Note 4.

Crystal structure generation is formulated as a flow-matching problem in the  latent space. Let $Z_{0}$ denote a latent representation sampled from the data distribution, and let $Z_{1}\sim\mathcal{N}(0,I)$ denote a sample from the base noise distribution which is a unit normal, with the same shape $Z_{0}$. For each time $t\sim\mathcal{U}(0,1)$, the linear interpolation path for the forward noising trajectory is defined as follows:
\[
Z_{t} = (1-t)Z_{0} + tZ_{1}.
\]

\par\medskip
\noindent\textbf{Denoising process.}
Inference begins by sampling an initial latent state 
$Z_{1} \sim \mathcal{N}(0,I)$ and integrating the learned reverse flow into the data manifold to obtain a final latent state $Z_{0}$. Unlike standard flow-matching models that approximate a single deterministic flow  velocity, GEM-x parametrizes the conditional distribution over denoising velocities $q_\theta(u \mid Z_t, t, c)$ as a Gaussian mixture:
\begin{equation}
q_\theta(u \mid Z_t, t, c) = \sum_{k=1}^{K} A_{\theta,k}(Z_t,t,c)\,
\mathcal{N}\!\left(u;\mu_{\theta,k}(Z_t,t,c),\Sigma_{\theta,k}(Z_t,t,c)\right).
\end{equation}
Here, $u$ denotes the latent velocity, $Z_t$ is the noisy latent state at time $t$, and $c$ denotes conditioning variables such as space group or target properties. The number of mixture components is $K$. For each component $k$, the neural network parameterized by $\theta$ predicts a mixture weight $A_{\theta,k}$, mean velocity $\mu_{\theta,k}$, and covariance $\Sigma_{\theta,k}$. The weights satisfy $\sum_{k=1}^{K}A_{\theta,k}=1$ and are obtained using a softmax over the predicted logits.

\par\medskip
\noindent\textbf{Loss Function.}
The objective is to minimize  the Kullback--Leibler divergence between the predicted velocity distribution \( q_\theta(u \mid Z_t) \) and the ground-truth velocity distribution \( p(u \mid Z_t) \). Since \( \log p(u \mid Z_t) \) does not depend on \( \theta \), the loss function reduces to:
\begin{equation}
\mathcal{L_{\theta}}
= \mathbb{E}_{t, Z_0, Z_t}
\bigl[ - \log q_\theta(u \mid Z_t) \bigr]
\end{equation}

The loss function can also be modified so that the network predicts the pure latent $Z_0$ instead of the velocity that points in its direction \citep{chen2025gaussian}. 
A reverse transition loss is used to prevent the gradient spikes caused by vanishing variances:
\begin{equation}
\mathcal{L}_{\theta}
= \mathbb{E}_{t,\, Z_{t-\Delta t},\, Z_t}
\bigl[ -\log q_\theta(Z_{t-\Delta t} \mid Z_t) \bigr]
\end{equation}
Supplementary Note 4 contains further information about the selection of the training objective.

\par\medskip
\noindent\textbf{Conditional guidance.}
The predicted distribution is conditioned on a target spacegroup for the generated material. To mitigate the tendency of standard classifier-free guidance methods \cite{ho2022classifierfree} to produce out of distribution samples \cite{chung2025cfg}, a probabilistic guidance scheme is employed to predict the conditional distribution $q_w(u \mid x_t, c)$. This formulation enhances conditional alignment while preserving the variance bounds of the original distribution. The guided distribution is defined as
\begin{equation}
q_w(u \mid Z_t,c)
= \frac{w(u)}{X} \, q_\theta(u \mid Z_t,c),
\end{equation}
where $w(u)$ is a guidance mask and $X$ is the normalization constant ensuring that $q_w$ integrates to one.

The guidance mask is constructed from the conditional and unconditional velocity distributions. We approximate these distributions by isotropic Gaussians 
$\mathcal{N}(\mu_c,s_c^2 I)$ and $\mathcal{N}(\mu_u,s_u^2 I)$, representing the conditional and unconditional denoising distribution respectively. The guidance mask is then:
\begin{equation}
w(u)
=
\frac{
\mathcal{N}\!\left(
u;\,
\mu_c + \tilde{w} s_c \Delta \mu_n,\,
(1-\tilde{w}^2)s_c^2 I
\right)
}{
\mathcal{N}(u;\mu_c,s_c^2 I)
},
\end{equation}
where $\tilde{w}\in[0,1)$ is the probabilistic guidance scale. The normalized deviation in the mean is:
\begin{equation}
\Delta \mu_n
=
\frac{\mu_c-\mu_u}{\|\mu_c-\mu_u\|/\sqrt{D}}.
\end{equation}
This construction biases probability mass toward denoising distributions aligned with the target space group while preserving variance bounds, reducing the off-manifold overshooting often observed with classifier-free guidance. The conditional guidance mechanism is further explained in Supplementary Note 4. 
\par\medskip
\noindent\textbf{Denoising Dynamics.}
Since the network  predicts a Gaussian mixture, the denoising trajectory may be traversed either via an SDE or an ODE. The conditional distribution is  sampled from in the SDE case, while in the ODE case, an expectation is taken over the conditional distribution. It can be shown that , within the limit of a large number of time steps, the two approaches converge in sample quality\cite{chen2025gaussian}.

\par\medskip
\noindent\textbf{Conditional mixture of experts.}
GEM-x employs a layer-wise, sparsely gated MoE architecture based on top-$k$ routing \citep{shazeer2017outrageously}, as illustrated in Figure~\ref{fig:arch_complete}. Each MoE layer comprises $n$ Transformer-based expert networks, of which only $k$ are activated to process the input tokens. This sparse activation increases the representational capacity of the network without a proportional increase in computational cost.

Expert selection is governed by a lightweight gating network conditioned on the diffusion timestep $t$ and the target space group $c$. The timestep and condition embeddings are passed through a multilayer perceptron (MLP) to generate routing logits, with additive Gaussian noise introduced to enable stochastic expert selection. The routing logits for expert $i$ are defined as
\begin{equation}
H_i(t,c) =
\operatorname{MLP}_i(t,c) + S\varepsilon_i,
\qquad
\varepsilon_i \sim \mathcal{N}(0,1),
\end{equation}
where $S$ controls the magnitude of the Gaussian perturbation. The resulting logits are converted into routing weights through softmax normalization followed by top-$k$ selection:
\begin{equation}
G(t,c) =
\operatorname{KeepTopK}
\left(
\operatorname{Softmax}(H(t,c)), k
\right).
\end{equation}

The top-$k$ operation retains the $k$ largest gating weights and sets the remaining weights to zero, producing a sparse gating mask that determines which experts are activated. Importantly, the gating network depends exclusively on $t$ and $c$, rather than the input representation $x$. This separates expert assignment from the representations processed by the experts, avoiding circular dependence in the routing mechanism.

Each selected expert is a Transformer-based subnetwork that processes the input representation $x$ conditioned on $t$ and $c$. Let $E_i(x,t,c)$ denote the output of the $i$-th expert and $G_i(t,c)$ its corresponding routing weight. The output of an MoE layer is computed as
\begin{equation}
y = \sum_{i=1}^{n} G_i(t,c) E_i(x,t,c),
\end{equation}
where $G_i(t,c)=0$ for all experts outside the selected top-$k$ set. Consequently, only the selected experts need to be evaluated, while the complete ensemble retains the representational capacity of all $n$ experts. Conditioning expert routing on $t$ and $c$ enables specialization across different stages of the denoising trajectory and spacegroups, allowing GEM-x to model complex latent representations efficiently.

\section*{Code Availability}
All scripts for GEM-x and examples of generated structures are available at \url{https://github.com/Mesbah-Lab-UCB/GEM-x.git}.

\section*{Conflict of Interest}
The authors declare no known competing conflict of interest.

\section*{Acknowledgements}
This work was supported in part by the U.S. Department of Energy, Office
of Science, Office of Fusion Energy Sciences, under award no. DESC0024472. The authors thank Alan Stoev for his valuable assistance in setting up the downstream materials evaluation pipeline, including machine-learning interatomic potential-based pre-relaxation and density functional theory calculations. This work used Anvil at Purdue University through allocation from the Advanced Cyberinfrastructure Coordination Ecosystem: Services \& Support (ACCESS) program, which is supported by U.S. National Science Foundation grants 2138259, 2138286, 2138307, 2137603, and 2138296. 
AI-assisted tools were used for grammatical corrections and language refinement. % The authors take full responsibility for the scientific content of the manuscript.

% ---------- References ----------

\bibliography{references}

\end{document}

\maketitle

\section*{Supplementary Note 1. Training and Implementation Details}
\label{sec:supp-implementation}

% ============================================================
% 3.1 Sampling, Conditioning, and Denoising
% ============================================================

\noindent \textbf{Sampling, conditioning, and denoising.}
The sampling configuration employed for crystal generation
with GEM-x is summarized in
Supplementary Table~\ref{tab:sampling-configuration}.
The configuration includes the guidance scale,
number of generated samples, sampling batch size,
number of denoising steps, conditioning variables,
and stochastic denoising parameters.

\begin{table}[!htbp]
    \centering
    \caption{
        Sampling, conditioning, and denoising
        configuration used for GEM-x.
    }
    \label{tab:sampling-configuration}
    \begin{tabular}{ll}
        \toprule
        \textbf{Parameter} & \textbf{Value} \\
        \midrule
        Classifier-free guidance scale & 0.09 \\
        Number of samples & 1,000 \\
        Sampling batch size & 100 \\
        Number of denoising steps & 256 \\
        Dataset-index conditioning & True \\
        Space-group conditioning & True \\
        Stochastic denoising & True \\
        Noise standard deviation & 1.0 \\
        \bottomrule
    \end{tabular}
\end{table}
\FloatBarrier

% ============================================================
% 3.2 Architectural Choices
% ============================================================

\noindent \textbf{Architectural choices.} The architectural specifications of the best-performing
GEM-x model are summarized in
Supplementary Table~\ref{tab:model-architecture}.
Each expert comprises two transformer layers,
with the first operating at the model dimension
and the second at the projection dimension.
The transformer output is subsequently processed
by a multilayer perceptron (MLP) with the
specified hidden dimension. The mixture-of-experts (MoE) module comprises
eight experts, of which the top two are selected
for each input through the gating mechanism.

\begin{table}[!htbp]
    \centering
    \caption{
        Architectural specifications of the
        best-performing GEM-x model.
    }
    \label{tab:model-architecture}
    \begin{tabular}{ll}
        \toprule
        \textbf{Component} & \textbf{Value} \\
        \midrule
        Autoencoder latent dimension & 8 \\
        Model dimension & 512 \\
        Projection dimension & 1024 \\
        Number of layers & 12 \\
        Number of attention heads & 8 \\
        MLP dimension & 1,536 \\
        Dropout probability & 0.1 \\
        Number of space groups & 230 \\
        Number of Gaussian components & 8 \\
        Number of experts & 8 \\
        Top-$k$ experts & 2 \\
        \bottomrule
    \end{tabular}
\end{table}

\FloatBarrier
\noindent \textbf{Optimization and learning-rate scheduling.} The model is optimized using Adam with cosine-annealing learning-rate
decay. The corresponding hyperparameters are reported in
Table~\ref{tab:optimization-hyperparameters}.

\begin{table}[!htbp]
    \centering
    \caption{Optimization and learning-rate scheduler hyperparameters.}
    \label{tab:optimization-hyperparameters}
    \begin{tabular}{ll}
        \toprule
        \textbf{Hyperparameter} & \textbf{Value} \\
        \midrule
        Optimizer                    & Adam \\
        Initial learning rate        & $1 \times 10^{-4}$ \\
        
        Learning-rate scheduler      & Cosine annealing \\
        Scheduler frequency          & 1 \\
        Scheduler period, $T_{\max}$ & 10,000 \\
        Minimum learning rate        & $1 \times 10^{-6}$ \\
        Load-balancing coefficient   & 6 \\
        Gradient-clipping value      & 10.0 \\
        Gradient-clipping algorithm  & Norm clipping \\
        \bottomrule
    \end{tabular}
\end{table}

\FloatBarrier
\noindent \textbf{Training configuration.} Table~\ref{tab:training-configuration} provides the remaining training,
validation, logging, and hardware settings. Training beyond 4250 epochs increases validity rates, but decreases uniqueness and novelty to a point that the total percentage of VUN structures decreases, and therefore further training is no longer conducive.

\begin{table}[!htbp]
    \centering
    \caption{Training and hardware configuration.}
    \label{tab:training-configuration}
    \begin{tabular}{ll}
        \toprule
        \textbf{Parameter} & \textbf{Value} \\
        \midrule
        Minimum number of epochs          & 1 \\
        Maximum number of epochs          & 4250 \\
        Training batch size               & 128 \\
        Validation batch size             & 128 \\
        Test batch size                   & 128 \\
        Accelerator                       & GPU \\
        Number of devices                 & 8 \\
        GPU type                          & A40 \\
        Validation frequency              & Every 250 epochs \\
        Metric-logging frequency          & Every training step \\
        
        \bottomrule
    \end{tabular}
\end{table}

\FloatBarrier

\section*{Supplementary Note 2. Downstream Analysis, Structure Screening and DFT Relaxation}
\label{sec:supp-downstream}

Generated crystal structures are screened and relaxed
using the following workflow.

\begin{enumerate}

    \item \textbf{Candidate generation.}
    An initial set of 1,000 candidate crystal structures
    is generated using GEM-x.

    \item \textbf{Initial structural filtering.}
    The generated structures are evaluated for validity,
    uniqueness, and novelty. Only structures satisfying
    all three criteria are retained for subsequent screening.

    \item \textbf{Machine-learning pre-relaxation.}
    As the generated structures are not necessarily located
    at local energy minima, they are pre-relaxed using the
    pretrained SevenNet machine-learning interatomic
    potential \cite{park2024sevennet}.
    Structural relaxation is performed until the maximum
    atomic force falls below
    \SI{0.05 }{ \electronvolt\per\angstrom}.
    The pre-relaxation calculations are constructed and
    managed using the force-field workflow components
    implemented in atomate2 \cite{ganose2025atomate2}.

    \item \textbf{Initial thermodynamic-stability screening.}
    Following pre-relaxation, the energy above the
    Materials Project GGA/GGA+$U$ convex hull,
    $E_{\mathrm{hull}}^{\mathrm{GGA}}$,
    is evaluated \cite{jain2013materialsproject}.
    Structures satisfying
    \begin{equation}
        E_{\mathrm{hull}}^{\mathrm{GGA}}
        \leq 0.15~\mathrm{eV\,atom^{-1}}
        \label{eq:initial-hull-screening}
    \end{equation}
    are selected for subsequent DFT relaxation.

    \item \textbf{Magnetic-moment initialization.}
    Before performing spin-polarized DFT calculations,
    site-resolved initial magnetic moments are predicted
    using the pretrained Crystal Hamiltonian Graph
    Neural Network (CHGNet) \cite{deng2023chgnet}.
    These predictions are used to initialize the
    magnetic moments of the candidate structures.

    \item \textbf{DFT relaxation and static calculation.}
    The selected structures are evaluated using the
    \texttt{MPMetaGGADoubleRelaxStaticMaker} workflow
    implemented in atomate2 \cite{ganose2025atomate2}.
    The workflow comprises three sequential stages:
    \begin{enumerate}
        \item an initial PBEsol pre-relaxation;
        \item a higher-accuracy $r^2$SCAN structural
        relaxation; and
        \item a final $r^2$SCAN static calculation
        on the relaxed structure.
    \end{enumerate}
    The energy obtained from the final static calculation
    is used for subsequent thermodynamic-stability analysis.

    \item \textbf{Final thermodynamic-stability evaluation.}
    The final DFT energies are evaluated relative to the
    Materials Project $r^2$SCAN
    convex hull
    \cite{jain2013materialsproject,kingsbury2022mixing}.
    The resulting energy above hull,
    $E_{\mathrm{hull}}$,
    is used to classify each structure as stable,
    metastable, or unstable.

    \item \textbf{Post-relaxation uniqueness and novelty checks.}
    Uniqueness and novelty are reassessed following
    structural relaxation. This additional evaluation
    accounts for initially distinct structures that may
    converge to the same local energy minimum during
    relaxation, as well as generated structures that
    may relax to configurations already present in the
    reference dataset. Duplicate and non-novel
    structures are excluded from the final set.

\end{enumerate}

\section*{Supplementary Note 3. Additional Results}
\label{sec:supp-experimental}

This note provides additional analyses of the crystal structures
generated by GEM-x and the routing behavior of its conditional
mixture-of-experts (MoE) architecture. We examine the distribution
of crystallographic space groups among generated structures and
characterize expert routing as a function of diffusion timestep
and crystallographic space-group conditioning.

% Required packages:
% \usepackage{graphicx}
% \usepackage{subcaption}
% \usepackage{placeins}

\begin{figure*}[t!]
    \centering

    \begin{subfigure}{0.32\textwidth}
        \centering
        \includegraphics[width=\linewidth]
        {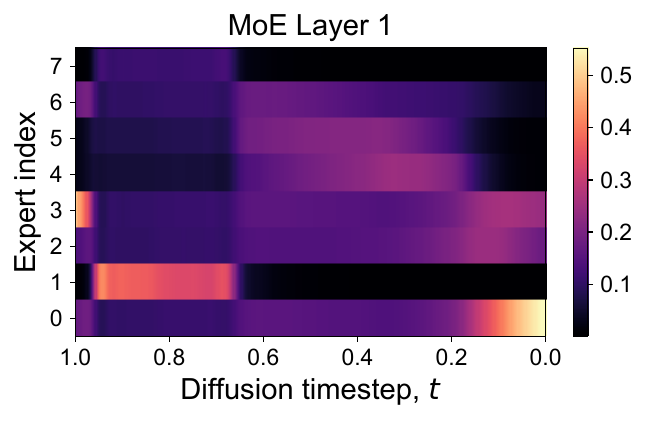}
    \end{subfigure}
    \hfill
    \begin{subfigure}{0.32\textwidth}
        \centering
        \includegraphics[width=\linewidth]
        {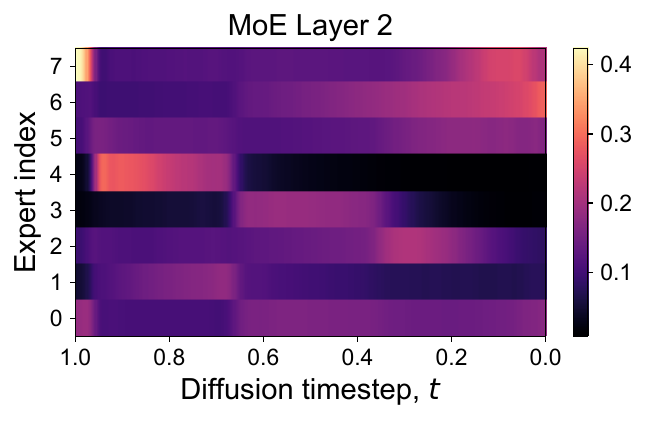}
    \end{subfigure}
    \hfill
    \begin{subfigure}{0.32\textwidth}
        \centering
        \includegraphics[width=\linewidth]
        {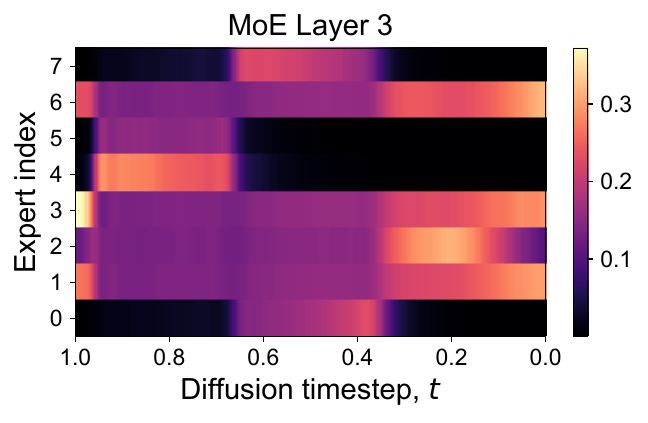}
    \end{subfigure}

    \medskip

    \begin{subfigure}{0.32\textwidth}
        \centering
        \includegraphics[width=\linewidth]
        {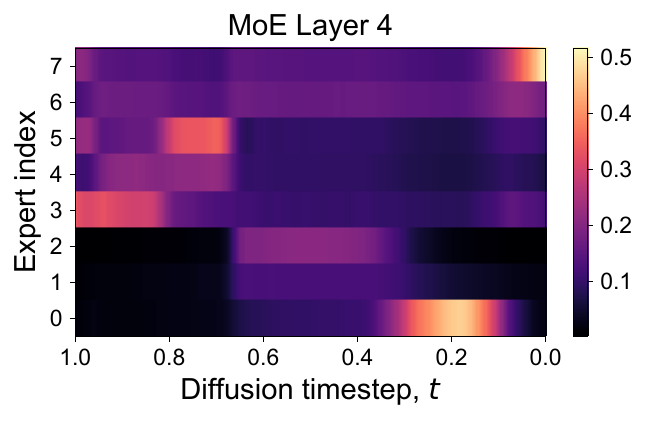}
    \end{subfigure}
    \hfill
    \begin{subfigure}{0.32\textwidth}
        \centering
        \includegraphics[width=\linewidth]
        {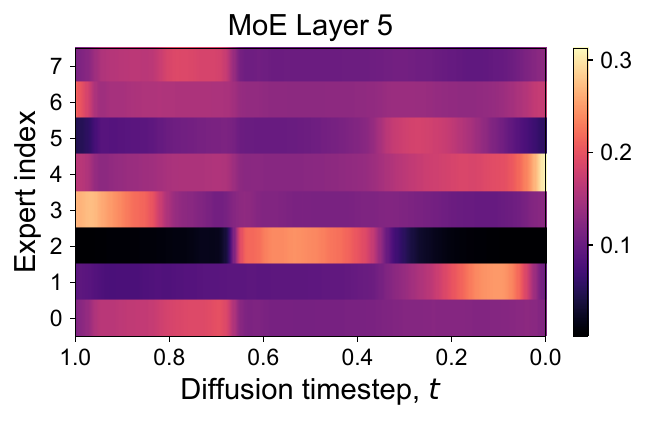}
    \end{subfigure}
    \hfill
    \begin{subfigure}{0.32\textwidth}
        \centering
        \includegraphics[width=\linewidth]
        {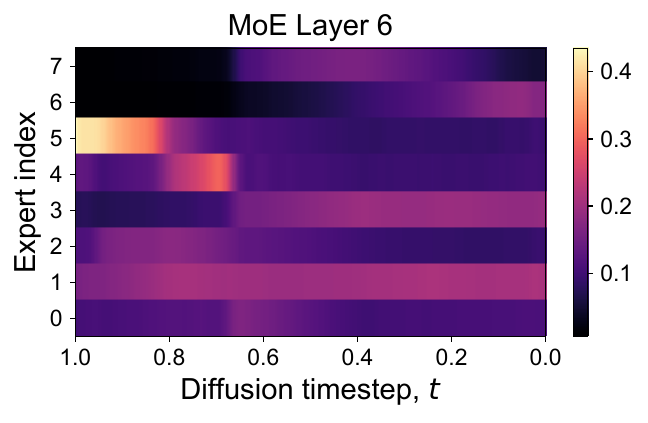}
    \end{subfigure}

    \medskip

    \begin{subfigure}{0.32\textwidth}
        \centering
        \includegraphics[width=\linewidth]
        {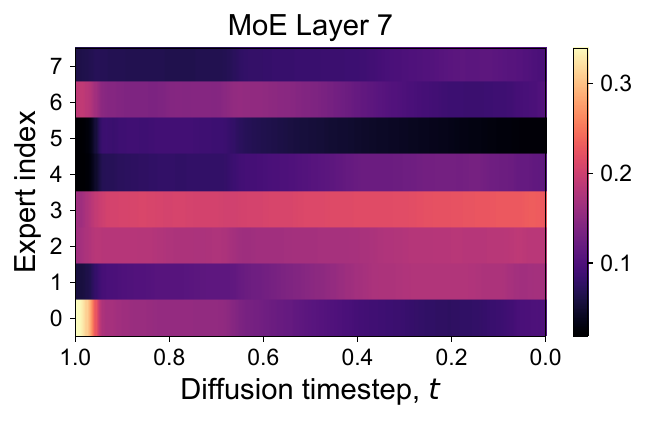}
    \end{subfigure}
    \hfill
    \begin{subfigure}{0.32\textwidth}
        \centering
        \includegraphics[width=\linewidth]
        {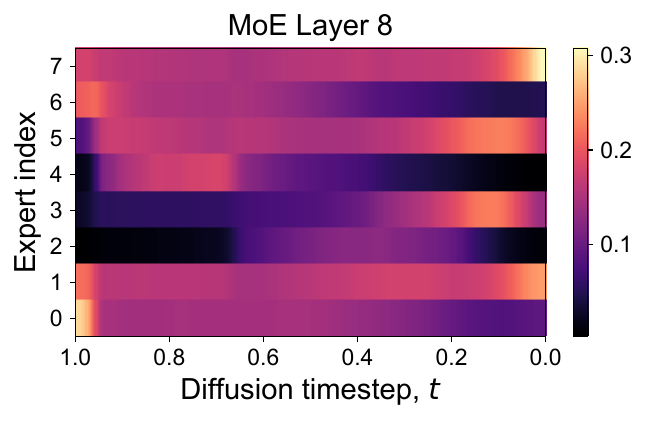}
    \end{subfigure}
    \hfill
    \begin{subfigure}{0.32\textwidth}
        \centering
        \includegraphics[width=\linewidth]
        {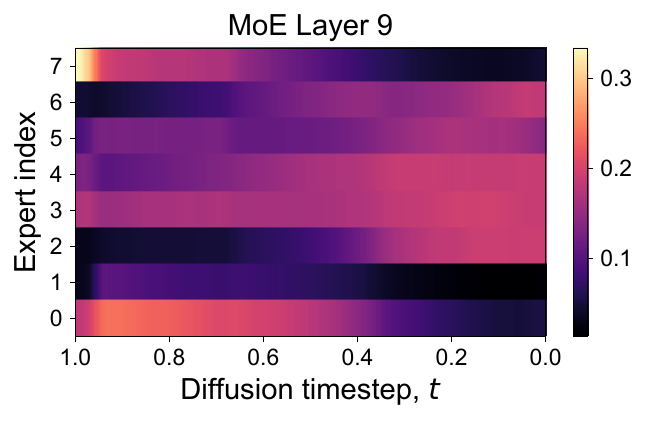}
    \end{subfigure}

    \medskip

    \begin{subfigure}{0.32\textwidth}
        \centering
        \includegraphics[width=\linewidth]
        {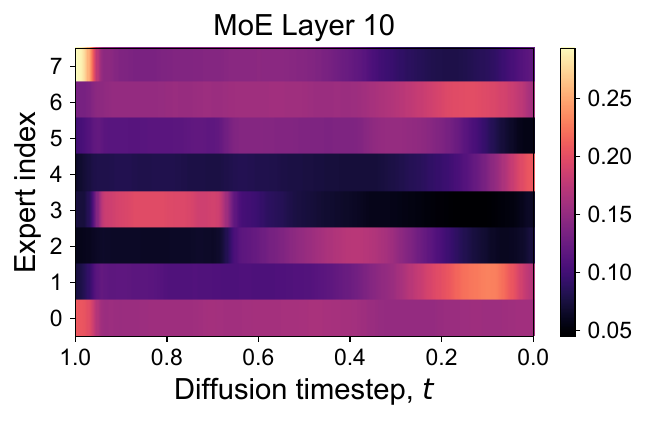}
    \end{subfigure}
    \hfill
    \begin{subfigure}{0.32\textwidth}
        \centering
        \includegraphics[width=\linewidth]
        {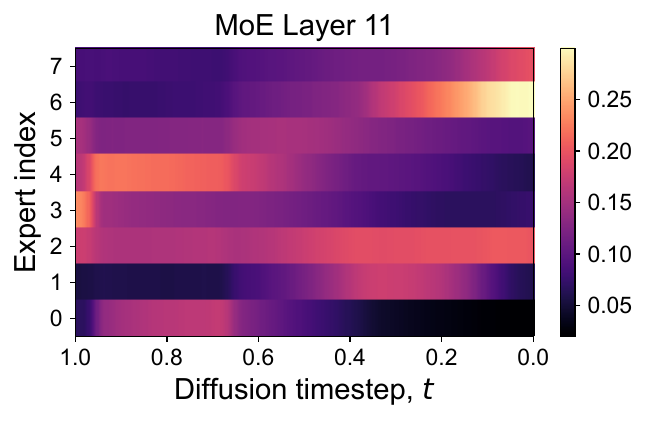}
    \end{subfigure}
    \hfill
    \begin{subfigure}{0.32\textwidth}
        \centering
        \includegraphics[width=\linewidth]
        {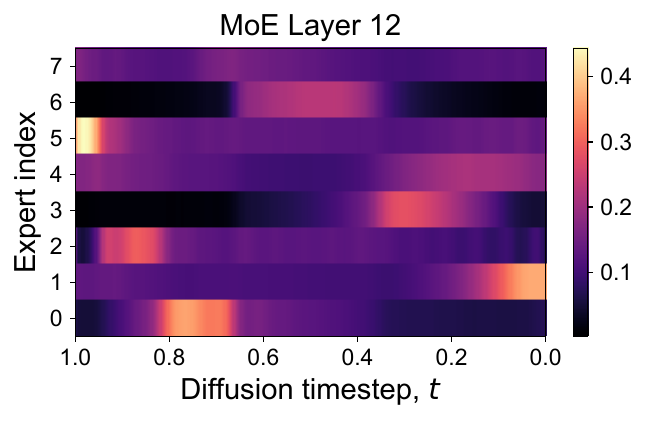}
    \end{subfigure}

    \caption{
        Timestep-dependent expert routing across the MoE layers.
        Each heatmap shows the output of the gating network, in terms of weights assigned to each expert, averaged over a batch of generated structures. Rows correspond to experts and columns correspond to diffusion
        timestep \(t\), with high-noise timesteps on the left and low-noise
        timesteps on the right. In each case, the number of called experts in sparse gating is 2.
    }
    \label{fig:moe-timestep-specialization}
\end{figure*}

\noindent \textbf{Timestep-dependent expert routing.} To examine how the conditional MoE architecture distributes
computation throughout the generative process, we analyze
the expert-routing weights as a function of diffusion timestep.
Supplementary Fig.~\ref{fig:moe-timestep-specialization}
presents the averaged gating-network outputs for each of
the twelve MoE layers. The heatmaps illustrate the relative contributions of
individual experts at different stages of denoising.
The timestep decreases from left to right, corresponding
to a progression from high-noise latent representations
toward low-noise representations.

\FloatBarrier

\begin{figure*}[!hbtp]
    \centering

    \begin{subfigure}{0.32\textwidth}
        \centering
        \includegraphics[width=\linewidth]
        {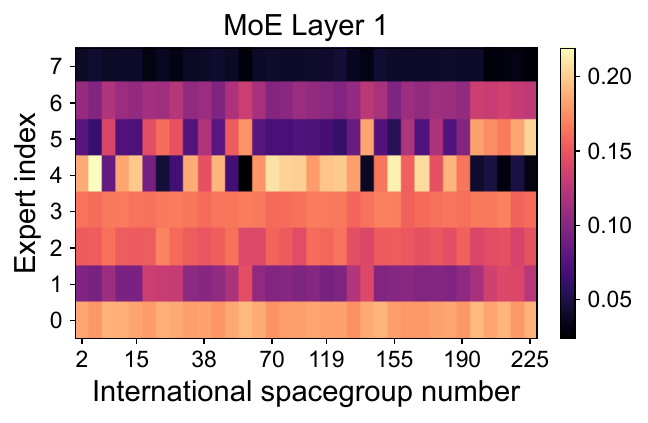}
    \end{subfigure}
    \hfill
    \begin{subfigure}{0.32\textwidth}
        \centering
        \includegraphics[width=\linewidth]
        {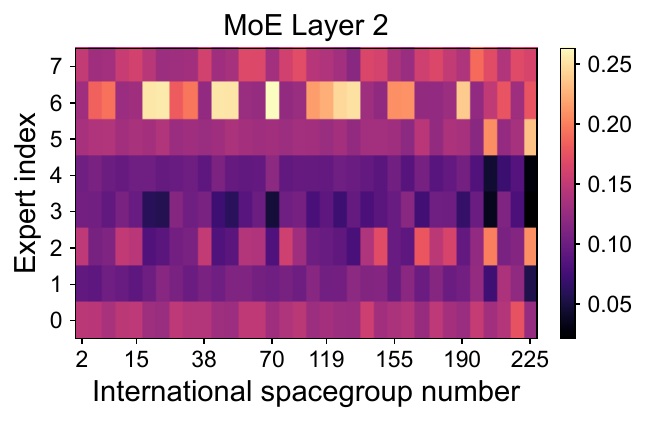}
    \end{subfigure}
    \hfill
    \begin{subfigure}{0.32\textwidth}
        \centering
        \includegraphics[width=\linewidth]
        {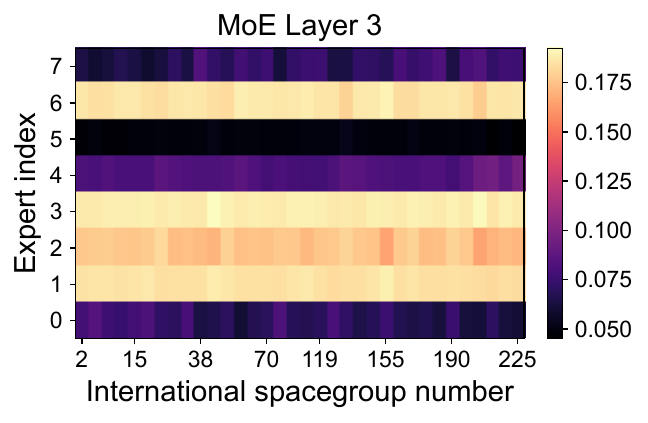}
    \end{subfigure}

    \medskip

    \begin{subfigure}{0.32\textwidth}
        \centering
        \includegraphics[width=\linewidth]
        {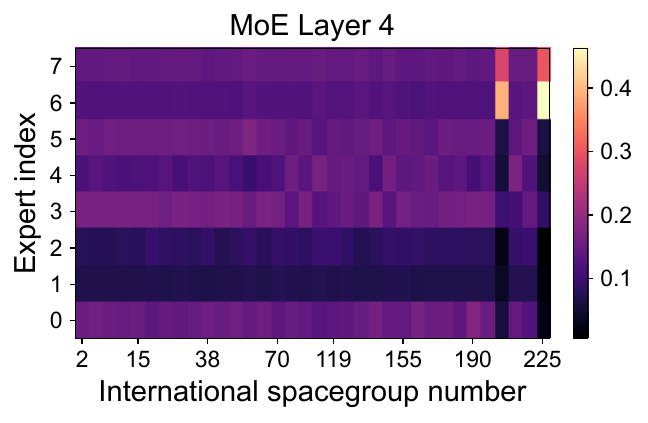}
    \end{subfigure}
    \hfill
    \begin{subfigure}{0.32\textwidth}
        \centering
        \includegraphics[width=\linewidth]
        {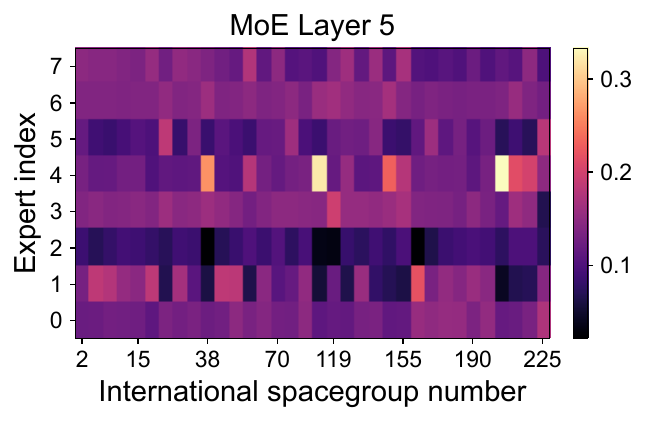}
    \end{subfigure}
    \hfill
    \begin{subfigure}{0.32\textwidth}
        \centering
        \includegraphics[width=\linewidth]
        {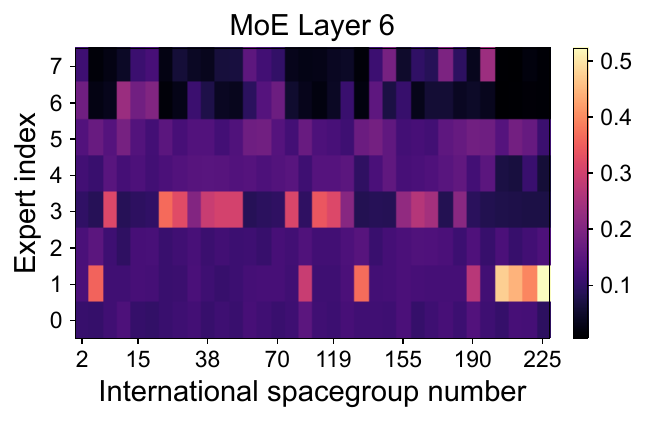}
    \end{subfigure}

    \medskip

    \begin{subfigure}{0.32\textwidth}
        \centering
        \includegraphics[width=\linewidth]
        {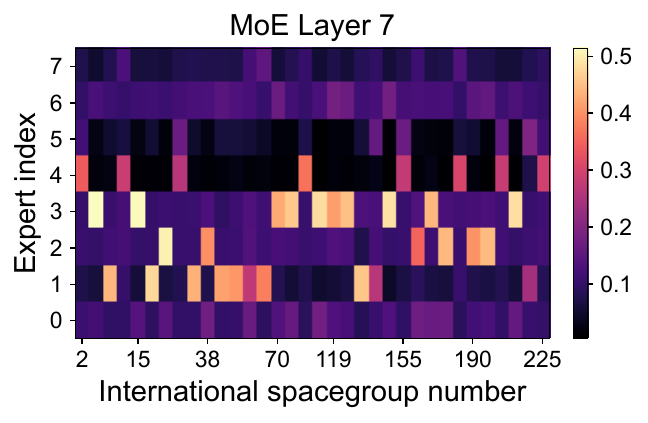}
    \end{subfigure}
    \hfill
    \begin{subfigure}{0.32\textwidth}
        \centering
        \includegraphics[width=\linewidth]
        {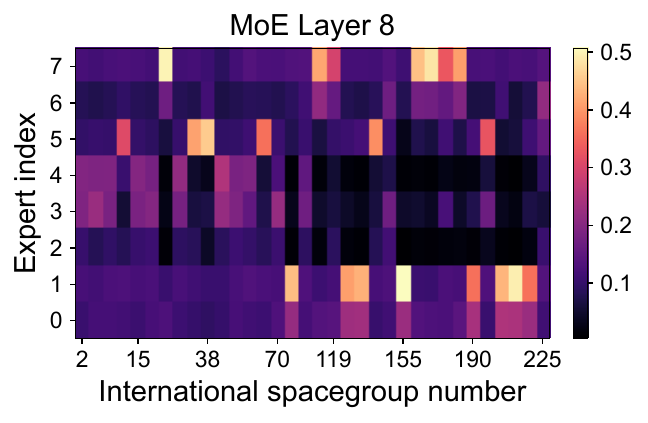}
    \end{subfigure}
    \hfill
    \begin{subfigure}{0.32\textwidth}
        \centering
        \includegraphics[width=\linewidth]
        {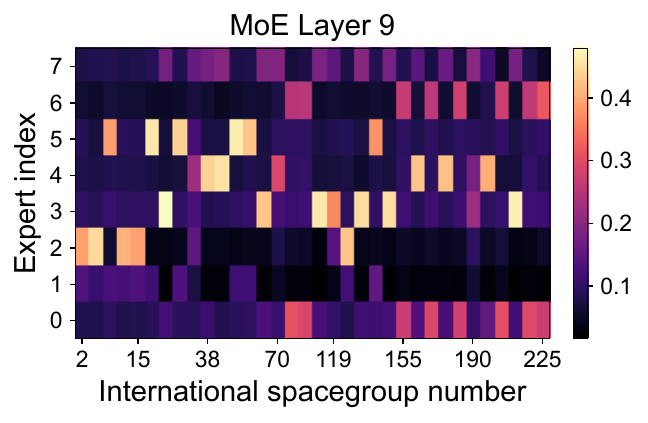}
    \end{subfigure}

    \medskip

    \begin{subfigure}{0.32\textwidth}
        \centering
        \includegraphics[width=\linewidth]
        {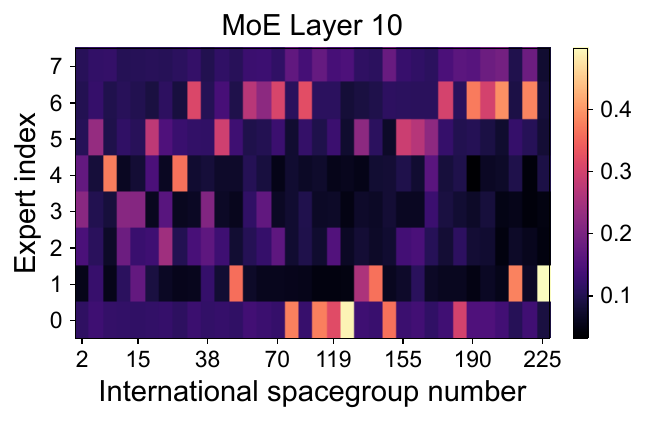}
    \end{subfigure}
    \hfill
    \begin{subfigure}{0.32\textwidth}
        \centering
        \includegraphics[width=\linewidth]
        {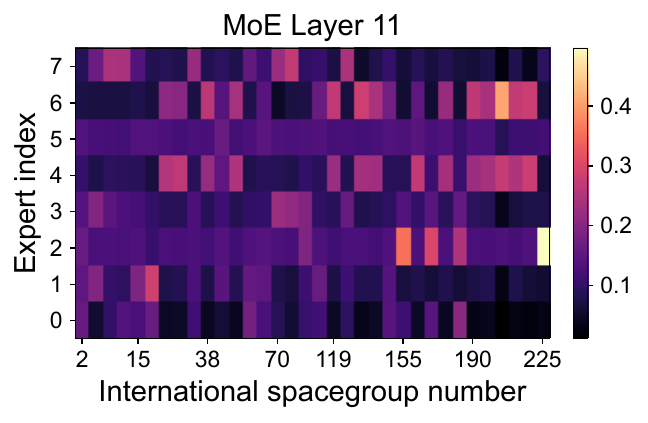}
    \end{subfigure}
    \hfill
    \begin{subfigure}{0.32\textwidth}
        \centering
        \includegraphics[width=\linewidth]
        {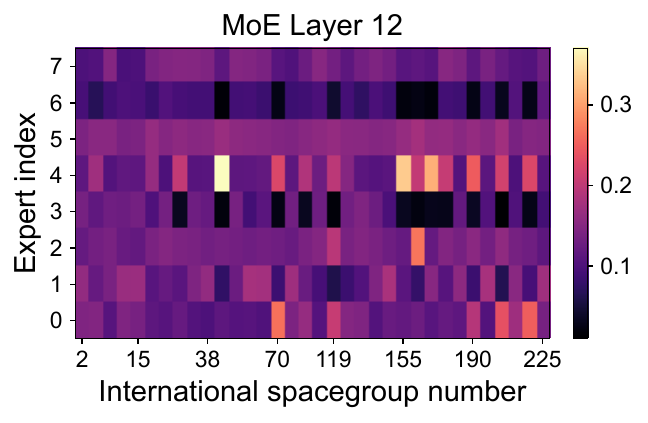}
    \end{subfigure}

    \caption{
        Space-group-dependent expert routing across the MoE layers.
        Columns correspond to the International space-group numbers and rows correspond to expert indices.  The color scale of each panel spans the
        activation range observed in that layer.
    }
    \label{fig:moe-spacegroup-specialization}
\end{figure*}

\noindent \textbf{Spacegroup-dependent expert routing.}
We further examine the dependence of expert routing on
crystallographic space-group conditioning.
Supplementary Fig.~\ref{fig:moe-spacegroup-specialization}
shows the routing weights associated with different
International space-group numbers across the twelve
MoE layers. These heatmaps provide a layer-wise characterization
of how the gating network distributes weights among
experts for different crystallographic symmetry conditions.
The visualizations complement the MoE analysis presented
in the main manuscript.

\FloatBarrier

\begin{figure}[!hbtp]
    \centering

    \begin{subfigure}[t]{1\textwidth}
        \centering
        \includegraphics[width=\linewidth]{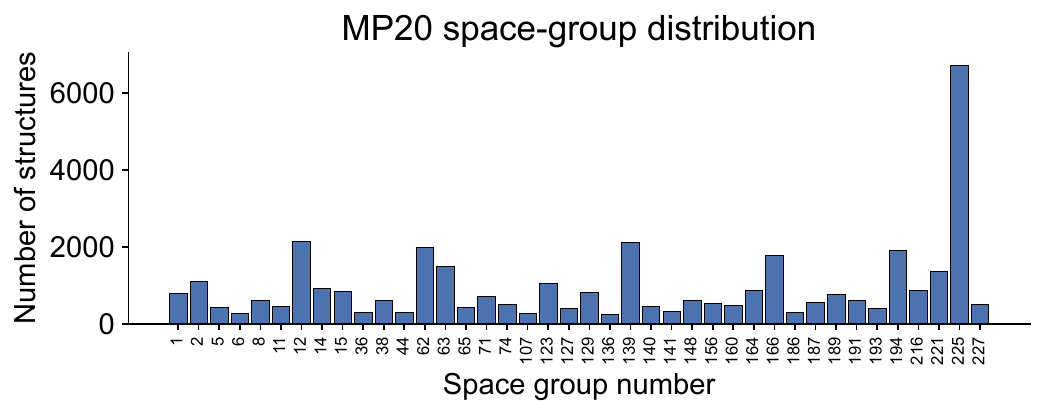}
        \caption{MP20 dataset}
        \label{fig:mp20-spacegroups}
    \end{subfigure}
    \hfill
    \begin{subfigure}[t]{1\textwidth}
        \centering
        \includegraphics[width=\linewidth]{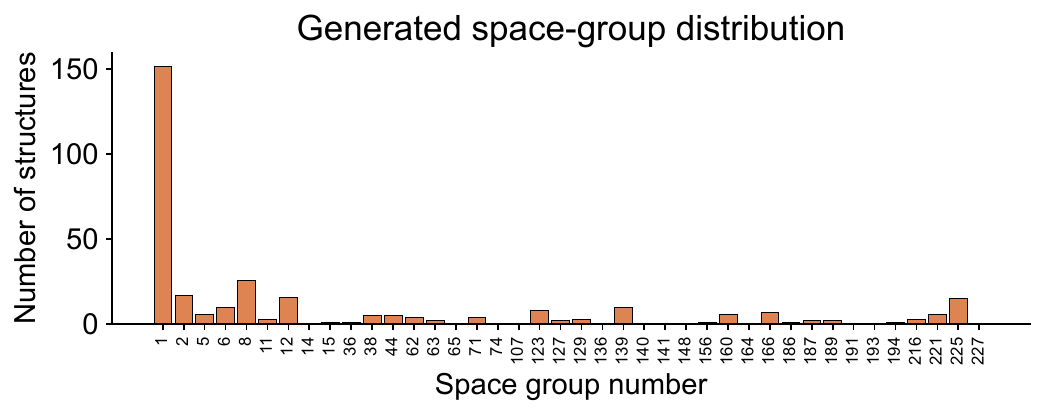}
        \caption{Generated samples}
        \label{fig:generated-spacegroups}
    \end{subfigure}

    \caption{
    Space-group distributions for the MP20  dataset and generated crystal structures which have been relaxed using SevenNet. After relaxation, these structures were metastable or stable,  unique and novel. Diffusion models have a tendency to over sample from the P1 spacegroup, which is the simplest spacegroup symmetry.
    }
    \label{fig:spacegroup-distributions}
\end{figure}

\noindent \textbf{Space-group distribution of generated crystals.}
Supplementary Fig.~\ref{fig:spacegroup-distributions} compares
the crystallographic space-group distributions of the MP20
dataset and the structures generated by GEM-x. The generated
structures shown have undergone relaxation using SevenNet and
satisfy the metastability or stability, uniqueness, and novelty
criteria employed in the screening procedure. The distributions provide additional characterization of the
crystallographic diversity of the generated structures,
including the relative representation of different space groups.

\clearpage

% ============================================================
% Supplementary Note 2: Mathematical Formulation of GEM-x
% ============================================================

\section*{Supplementary Note 4. Mathematical Formulation of GEM-x}
\label{sec:supp-methods}

\noindent \textbf{Unified atomic representations.} GEM-x adopts the unified atom-wise representation introduced
by the All-Atom Diffusion Transformer (ADiT)
\citep{joshi2025allatom}, which enables both periodic crystals
and non-periodic molecules to be represented in a shared
continuous latent space. This framework employs a variational
autoencoder (VAE) to map atomic structures, characterized by
categorical variables such as atomic species, and continuous
variables such as atomic positions, into continuous latent
embeddings. We use the VAE framework of ADiT to construct
the latent representation space for Gaussian-mixture flow
matching in GEM-x.

\emph{Unified representation of three-dimensional atomic systems.} An atomic system comprising $N$ atoms is described by its
atomic species $A$, Cartesian coordinates $X$, fractional
coordinates $F$, and lattice vectors $L$, defined as
\begin{equation}
\begin{aligned}
A &= \{a_i\}_{i=1}^{N} \in \mathbb{Z}^{1\times N},
&\qquad
X &= \{x_i\}_{i=1}^{N} \in \mathbb{R}^{3\times N},\\
F &= \{f_i\}_{i=1}^{N} \in [0,1)^{3\times N},
&\qquad
L &= [l_1,l_2,l_3] \in \mathbb{R}^{3\times 3}.
\end{aligned}
\label{eq:atomic-representation}
\end{equation}

For periodic crystals, the structure is fully specified
by $(A,F,L)$, with Cartesian coordinates recovered through
$X=LF$. For non-periodic molecules, the structure is
specified by $(A,X)$, while fractional coordinates and
lattice parameters are assigned a null value $\phi$.

\emph{Variational autoencoder architecture.} The VAE maps atomic structures into continuous per-atom
latent embeddings. Given the relevant structural variables,
the encoder $E$ produces
\begin{equation}
Z = E(A,X,F), \qquad
Z = \{z_i\}_{i=1}^{N} \in \mathbb{R}^{d\times N},
\label{eq:vae-encoder}
\end{equation}
where $d$ denotes the latent dimensionality, and N=20 denotes the maximum number of atoms in a crystal in MP20 dataset . The decoder
$D$ reconstructs the corresponding atomic representation,
\begin{equation}
(A',X',F',L') = D(Z).
\label{eq:vae-decoder}
\end{equation}

For periodic crystals, the reconstructed variables
$(A',F',L')$ define the periodic unit cell, whereas
for non-periodic molecules, $(A',X')$ define the
molecular structure.

\emph{Reconstruction objective.} The VAE is trained by minimizing a reconstruction loss
that quantifies the discrepancy between the reconstructed
atomic representation and the corresponding input.
A channel-wise Kullback--Leibler (KL) divergence penalty
is additionally included to regularize the latent
distribution toward a standard normal prior. For periodic crystals, the reconstruction objective
involves atomic species, fractional coordinates, and
lattice parameters. For non-periodic molecules, it
involves atomic species and Cartesian coordinates.
Further details of the VAE architecture and training
procedure are provided in Ref.~\citep{joshi2025allatom}.

\noindent \textbf{Latent flow matching.} Flow-matching generative models
\citep{lipman2023flowmatchinggenerativemodeling} transform a base distribution
into a target distribution by learning a time-dependent
vector field along an interpolation path between the
two distributions. In the latent space, this vector field
describes the flow velocity that governs the transformation
from noisy latent representations toward the latent
representations of the data.

Let $Z_{0}$ denote the clean latent representation
obtained from the VAE encoder and
$Z_{1} \sim \mathcal{N}(0,I)$ denote a sample from
the standard normal distribution. The interpolation
between these representations is defined as
\begin{equation}
Z_{t} = (1-t)Z_{0} + tZ_{1},
\qquad t \sim \mathcal{U}(0,1).
\label{eq:latent-interpolation}
\end{equation}

The corresponding ground-truth conditional vector
field is given by
\begin{equation}
u_t\left(Z_{t} \mid Z_{1}\right)
=
\frac{Z_{1}-Z_{t}}{1-t}.
\label{eq:flow-velocity}
\end{equation}

A denoising network $F$ predicts the noise endpoint
$Z'^{(1)}$ from the interpolated latent representation,
diffusion timestep, and conditioning variable:
\begin{equation}
Z'_{1} = F\left(Z_{t},t,c\right),
\label{eq:flow-denoiser}
\end{equation}
where $c$ is a binary label distinguishing periodic
and non-periodic atomic systems. The flow-matching
training objective is expressed as
\begin{equation}
\mathcal{L}_{\mathrm{fm}}
=
\frac{1}{N}
\sum_{i=1}^{N}
\frac{
\left\|Z^{(i)}_{1}-Z'^{(i)}_{1}\right\|^2
}{1-t}.
\label{eq:flow-matching-loss}
\end{equation}

\emph{Limitations of conventional flow matching.} The performance of flow-matching generative models
is influenced by several sources of error and
limitations:

\begin{enumerate}
    \item \textit{Discretization error:}
    Numerical integration of the learned ordinary
    or stochastic differential equations (ODEs/SDEs)
    introduces discretization errors.

    \item \textit{Flow-field approximation error:}
    Inaccuracies in the predicted vector field
    $u_\theta(Z_t,c)$ can arise from model underfitting.

    \item \textit{Unimodal velocity prediction:}
    A unimodal denoising velocity cannot explicitly
    represent multiple plausible trajectory endpoints
    from the same latent state.

    \item \textit{Limited sampling stochasticity:}
    When stochasticity is restricted to the initial
    sample from the base distribution, diversity
    in the generated samples may be limited.
\end{enumerate}

Reducing the integration step size can mitigate
discretization error, but increases the number of
function evaluations (NFE). Classifier-free guidance
(CFG) is commonly employed to improve conditional
predictions by combining conditional and unconditional
vector fields:
\begin{equation}
u_{\theta}^{\mathrm{CFG}}(Z_t,c)
=
w\,u_\theta(Z_t,c)
+
(1-w)\,u_\theta(Z_t),
\qquad w\in[1,\infty).
\label{eq:classifier-free-guidance}
\end{equation}

Although increasing the guidance weight $w$ can
improve alignment with the target condition,
excessive guidance may lead to out-of-distribution
latent representations.

\noindent \textbf{Forward noising process.} The forward noising process transforms a  latent
representation $Z$ into a noisy latent representation $Z_t$
according to
\begin{equation}
Z_t = \alpha_t Z + \sigma_t \varepsilon,
\qquad
\varepsilon \sim \mathcal{N}(0,I),
\label{eq:forward-noising}
\end{equation}
where $\alpha_t$ and $\sigma_t$ are predefined, time-dependent
noise-schedule coefficients.

The marginal distribution of the noisy latent variables
$p(Z_t)$ is obtained by convolving the clean latent
distribution $p(Z)$ with a Gaussian kernel. The corresponding
conditional distribution is given by
\begin{equation}
p(Z_t \mid Z_0)
=
\mathcal{N}
\left(
Z_t;\alpha_t Z_0,\sigma_t^2 I
\right).
\label{eq:forward-transition}
\end{equation}

The noise schedule satisfies the boundary conditions
\begin{equation}
Z_0 = Z,
\qquad
Z_T \approx \varepsilon.
\label{eq:noising-boundary}
\end{equation}

For the linear noising schedule, the terminal time
and schedule coefficients are defined as
\begin{equation}
T=1,
\qquad
\alpha_t=1-t,
\qquad
\sigma_t=t.
\label{eq:linear-noise-schedule}
\end{equation}

\noindent \textbf{Reverse denoising process and loss function.}
The denoising flow velocity is modeled using a Gaussian
mixture distribution,
\begin{equation}
q_\theta(u \mid Z_t)
=
\sum_{k=1}^{K}
A_{k,\theta}\,\mathcal{N}(u;\mu_{k,\theta},\Sigma_{k,\theta}),
\label{eq:gaussian-mixture}
\end{equation}
where $K$ denotes the number of mixture components,
$A_k$ is the mixture weight, and $\mu_k$ and $\Sigma_k$
denote the mean and covariance of the $k$-th component,
respectively.

The negative log-likelihood objective is defined as
\begin{equation}
\mathcal{L}
=
\mathbb{E}_{t,Z_0,Z_t}
\left[
-\log q_\theta(u\mid Z_t)
\right].
\label{eq:mixture-nll}
\end{equation}

Using an isotropic covariance parameterization,
$\Sigma_k=s^2I$, the objective can be expressed as
\begin{equation}
\begin{aligned}
\mathcal{L}
=
\mathbb{E}_{t,Z_0,Z_t}
\Bigg[
-\log
\sum_{k=1}^{K}
\exp\Bigg(
&-\frac{\|u-\mu_k(Z_t,t)\|^2}{2s^2}\\
&-D\log s
+\log A_k(Z_t,t)
\Bigg)
\Bigg],
\end{aligned}
\label{eq:mixture-nll-expanded}
\end{equation}
where $D$ denotes the dimensionality of the predicted
velocity and $s>0$ is a learned scalar shared across
mixture components. The expression omits additive
constants independent of the model parameters.

Although the isotropic covariance formulation reduces
the complexity of the mixture model, it does not
enforce a positive lower bound on the learned variance.
Consequently, the variance may approach zero during
optimization, leading to numerical instability and
large gradients.

To improve training stability, we instead adopt a
reverse-transition likelihood objective \cite{chen2025gaussian},
\begin{equation}
\mathcal{L}_{\mathrm{trans}}
=
\mathbb{E}_{t,Z_{t-\Delta t},Z_t}
\left[
-\log q_\theta
\left(
Z_{t-\Delta t}\mid Z_t
\right)
\right].
\label{eq:transition-loss}
\end{equation}

For $\Delta t<t$, the reverse-transition distribution
maintains a lower bound on its variance, which
improves numerical stability during training.

The transition time is parameterized as
\begin{equation}
t-\Delta t=\lambda t,
\qquad
\lambda\in(0,1],
\label{eq:transition-ratio}
\end{equation}
where $\lambda$ is the transition-ratio hyperparameter.

\noindent \textbf{Conditional guidance.}
The guided conditional distribution
$q_w(u \mid Z_t,c)$ is defined as
\begin{equation}
q_w(u \mid Z_t,c)
=
\frac{w(u)}{X}
q_\theta(u \mid Z_t,c),
\label{eq:guided-distribution}
\end{equation}
where $X$ denotes the normalization constant and
$w(u)$ is a Gaussian guidance mask. The guidance mask is constructed by approximating
the conditional and unconditional Gaussian-mixture
distributions with isotropic Gaussians,
$\mathcal{N}(\mu_c,s_c^2I)$ and
$\mathcal{N}(\mu_u,s_u^2I)$, respectively.
These approximations match the means and total
variances of the corresponding mixture distributions.

The guidance mask is given by
\begin{equation}
w(u)
=
\frac{
\mathcal{N}\!\left(
u;\,
\mu_c+\tilde{w}s_c\Delta\mu_n,\,
(1-\tilde{w}^2)s_c^2I
\right)
}{
\mathcal{N}\!\left(
u;\mu_c,s_c^2I
\right)
},
\label{eq:guidance-mask}
\end{equation}
where $\tilde{w}\in[0,1)$ denotes the
probabilistic guidance scale. The normalized
difference between the conditional and
unconditional means is defined as
\begin{equation}
\Delta\mu_n
=
\frac{\mu_c-\mu_u}
{\|\mu_c-\mu_u\|/\sqrt{D}},
\label{eq:normalized-mean-difference}
\end{equation}
where $D$ denotes the dimensionality of
the predicted variable $u$.

\noindent \textbf{Denoising dynamics.} Here we discuss denoising dynamics.

\emph{GM-SDE solver.} Given the predicted Gaussian-mixture distribution
$q_\theta(Z_0\mid Z_t)$, the reverse-transition
distribution can be derived analytically as
\begin{equation}
\begin{aligned}
q_\theta(Z_{t-\Delta t}\mid Z_t)
=
\sum_{k=1}^{K} A_k\,
\mathcal{N}\!\Bigl(
&Z_{t-\Delta t};\\
&c_1Z_t+c_2\mu_k,\,
(c_3+c_2^2s^2)I
\Bigr),
\end{aligned}
\label{eq:gm-sde-transition}
\end{equation}
where the coefficients $c_1$, $c_2$, and $c_3$
depend only on the noise schedule. Sampling from
this distribution yields a Gaussian-mixture
approximation to the reverse-time stochastic
differential equation (SDE).

\emph{GM-ODE solver.} The Gaussian-mixture formulation also enables
an ordinary differential equation (ODE) solver
that analytically derives the velocity field
at intermediate times without requiring
additional network evaluations.

Given $q_\theta(Z_0\mid Z_t)$, the denoising
distribution at any intermediate time $\tau<t$
is expressed as
\begin{equation}
\hat{q}(Z_0\mid Z_\tau)
=
\frac{
p(Z_\tau\mid Z_0)
}{
Z\,p(Z_t\mid Z_0)
}
q_\theta(Z_0\mid Z_t),
\label{eq:gm-ode-distribution}
\end{equation}
which remains a Gaussian mixture.
The corresponding velocity field is obtained
through reparameterization and computation
of the mixture mean, enabling low-error
integration over intermediate substeps.

\noindent \textbf{Conditional mixture of experts.} Let $G(t,c)$ denote the output of the gating network,
and let $E_i(x,t,c)$ denote the output of the $i$-th
expert network for an input representation $x$,
diffusion timestep $t$, and conditioning variable $c$.
The output $y$ of the mixture-of-experts (MoE)
module is computed as
\begin{equation}
y
=
\sum_{i=1}^{n}
G_i(t,c)\,E_i(x,t,c),
\label{eq:moe-output}
\end{equation}
where $n$ denotes the total number of experts
and $G_i(t,c)$ is the routing weight assigned
to expert $i$.

The gating weights are determined by a
top-$k$ selection operation applied to the
softmax-normalized routing logits:
\begin{equation}
G(t,c)
=
\operatorname{KeepTopK}
\left(
\operatorname{Softmax}(H(t,c)),k
\right),
\label{eq:moe-gating}
\end{equation}
where the routing logits are defined as
\begin{equation}
H_i(t,c)
=
\operatorname{MLP}_i(t,c)
+
S\,\varepsilon_i,
\qquad
\varepsilon_i\sim\mathcal{N}(0,1).
\label{eq:moe-routing-logits}
\end{equation}
The parameter $S$ controls the magnitude
of the Gaussian noise added to the routing
logits. The top-$k$ operation retains only
the selected expert weights and sets
the remaining weights to zero.
Consequently, experts for which
$G_i(t,c)=0$ need not be evaluated,
enabling sparse expert computation.

\bibliography{supp_bib}